\documentclass[twoside,11pt]{article}

\usepackage{jmlr2e}

\usepackage{natbib}

\usepackage{natbib}

\usepackage{algorithm}
\usepackage{algorithmic}

\usepackage{amssymb}
\usepackage{amsmath}
\usepackage{multirow}
\usepackage{mathrsfs}
\usepackage{bbm}
\usepackage{bm}
\usepackage{eufrak}
\usepackage{bookmark}
\usepackage{mathtools}

\ShortHeadings{Discriminative Similarity for Clustering and Semi-Supervised Learning}{Y. Yang et al.}
\firstpageno{1}

\newcommand{\cA}{\mathcal{A}}

\newcommand{\cH}{\mathcal{H}}

\newcommand{\cL}{\mathcal{L}}

\newcommand{\cfrakR}{\mathfrak{R}} 

\newcommand{\cS}{\mathcal{S}}

\newcommand{\cX}{\mathcal{X}}
\newcommand{\cY}{\mathcal{Y}}

\newcommand{\cP}{\mathcal{P}}

\newcommand{\bzero}{\bm{0}}
\newcommand{\bone}{\bm{1}}
\newcommand{\balpha}{\bm{\alpha}}

\newcommand{\bt}{\mathbf{t}}
\newcommand{\bx}{\mathbf{x}}
\newcommand{\by}{\mathbf{y}}
\newcommand{\bl}{\ell}

\newcommand{\bw}{\mathbf{w}}

\newcommand{\bD}{\mathbf{D}}

\newcommand{\bY}{\mathbf{Y}}

\newcommand{\bS}{\mathbf{S}}

\newcommand{\bF}{{\mathbf{F}}}

\newcommand{\bI}{{\mathbf{I}}}
\newcommand{\bL}{{\mathbf{L}}}
\newcommand{\bK}{{\mathbf{K}}}

\newcommand{\N}{{\rm I}\kern-0.18em{\rm N}}

\newcommand{\R}{{\rm I}\kern-0.18em{\rm R}}
\newcommand{\h}{{\rm I}\kern-0.18em{\rm H}}
\newcommand{\K}{{\rm I}\kern-0.18em{\rm K}}
\newcommand{\p}{{\rm I}\kern-0.18em{\rm P}}
\newcommand{\E}{{\rm I}\kern-0.18em{\rm E}}
\newcommand{\Z}{{\rm Z}\kern-0.18em{\rm Z}}

\newcommand{\1}{{\rm 1}\kern-0.25em{\rm I}}

\newcommand{\pn}{\p_{\kern-0.25em n}}
\newcommand{\pnm}{\p_{\kern-0.25em n,m}}
\newcommand{\psubm}{\p_{\kern-0.25em m}}

\newcommand{\BigO}[1]{{\operatorname{O}}}

\DeclareMathOperator*{\argmax}{arg\,max}

\newtheorem{MyDefinition}{Definition}
\newtheorem{MyLemma}{Lemma}
\newtheorem{MyTheorem}{Theorem}

\newtheorem{MyRemark}{Remark}

\begin{document}

\title{Discriminative Similarity for Clustering and Semi-Supervised Learning}

\author{\name Yingzhen Yang \email superyyzg@gmail.com \\
       \addr Snap Research, USA\\
       \AND
       \name Feng Liang \email liangf@illinois.edu \\
       \addr Department of Statistics, University of Illinois at Urbana-Champaign, USA \\
       \AND
       Nebojsa Jojic \email jojic@microsoft.com \\
       Microsoft Research, USA\\
       \AND
       Shuicheng Yan \email eleyans@nus.edu.sg \\
       \addr 360 AI Institute\&ECE at National University of Singapore\\
       \AND
       \name Jiashi Feng \email elefjia@nus.edu.sg \\
       \addr Department of ECE, National University of Singapore, Singapore \\
       \AND
       Thomas S. Huang \email t-huang1@illinois.edu \\
       \addr Beckman Institute, University of Illinois at Urbana-Champaign, USA\\
       }

\editor{}

\maketitle

\begin{abstract}
Similarity-based clustering and semi-supervised learning methods separate the data into clusters or classes according to the pairwise similarity between the data, and the pairwise similarity is crucial for their performance. In this paper, we propose a novel discriminative similarity learning framework which learns discriminative similarity for either data clustering or semi-supervised learning. The proposed framework learns classifier from each hypothetical labeling, and searches for the optimal labeling by minimizing the generalization error of the learned classifiers associated with the hypothetical labeling. Kernel classifier is employed in our framework. By generalization analysis via Rademacher complexity, the generalization error bound for the kernel classifier learned from hypothetical labeling is expressed as the sum of pairwise similarity between the data from different classes, parameterized by the weights of the kernel classifier. Such pairwise similarity serves as the discriminative similarity for the purpose of clustering and semi-supervised learning, and discriminative similarity with similar form can also be induced by the integrated squared error bound for kernel density classification. Based on the discriminative similarity induced by the kernel classifier, we propose new clustering and semi-supervised learning methods.
\end{abstract}


\section{Introduction}
Similarity-based clustering and semi-supervised learning methods segment the data based on the similarity measure between the data points.
Regarding to similarity-based data clustering,  spectral clustering \citep{Ng01} identifies clusters of complex shapes lying on some low dimensional manifolds by normalized graph Laplacian from a data similarity matrix. Pairwise clustering method \citep{ShentalZHW03} uses message-passing algorithm to infer the cluster labels in a pairwise undirected graphical model with the pairwise potential function constructed from a similarity matrix. K-means \citep{HartiganW79} searches for data clusters by a local minimum of sum of within-cluster dissimilarities. 

The representative similarity-based semi-supervised learning algorithm is label propagation \citep{ZhuGL03,ZhouBLWS03}, which is effective and widely used. With a predefined similarity graph, it determines the labels of unlabeled data by the minimization of the objective function over the similarity graph defined as sum of the product of pairwise similarity and the squared label difference. Therefore, label propagation encourages local smoothness of the labels according to the edge weight of the similarity graph. The typical label propagation algorithm \citep{ZhuGL03} renders a harmonic solution which can also be interpreted by random walks from unlabeled data to the labeled data.
The success of similarity-based learning method highly depends on the underlying pairwise similarity over the data, which in most cases are constructed empirically, e.g. by Gaussian kernel or the K-Nearest-Neighbor (KNN) graph. In this paper, we present a discriminative similarity learning framework for clustering and semi-supervised learning wherein the discriminative similarity is derived as the generalization error bound for the kernel classifier learned from hypothetical labeling. When the popular Support Vector Machines (SVMs) is used in this framework, unsupervised SVM \citep{XuNLS04} is deduced for the unsupervised case, and Semi-Supervised or Transductive SVMs \citep{Vapnik1998,Joachims1999,Chapelle2008} is deduced for the semi-supervised learning case. A kernel classifier motivated by similarity learning \citep{Balcan08,CortesMR13} is used in our framework. By generalization analysis via Rademacher complexity, the generalization error bound for the kernel classifier learned from hypothetical labeling is expressed as the sum of pairwise similarity between the data from different classes. Such pairwise similarity, parameterized by the weights of the learned kernel classifier, serves as the discriminative similarity induced by this generalization bound for clustering and semi-supervised learning. \textit{Although similarity is often used to quantify the local affinity between the data, the term ``discriminative similarity" here means the similarity to be learned that improve the discriminative capability of some classification method such as the mentioned kernel classifier.} Moreover, we prove that discriminative similarity with the same form can also be induced by the error bound for the integrated squared error of kernel density classification.


Our discriminative similarity learning framework is related to a class of discriminative clustering methods which classify unlabeled data by various measures on the discriminative unsupervised classifiers, and the measures include generalization error \citep{XuNLS04} or the entropy of the posterior distribution of the label \citep{GomesKP10}. Discriminative clustering methods \citep{XuNLS04} predict the labels of unlabeled data by minimizing the generalization error bound for the unsupervised classifier. \citep{XuNLS04} proposes Unsupervised SVMs which learns a binary classifier to partition unlabeled data with the maximum margin between different clusters. The theoretical properties of unsupervised SVMs are analyzed in \citep{Karnin12}. \citep{GomesKP10} learns the kernel logistic regression classifier regularized by the entropy of the posterior distribution of the class label.

The paper is organized as follows. We introduce the formulation of the discriminative similarity learning framework in Section~\ref{sec:discriminative-similarity-framework}, and then derive the generalization error bound for the kernel classifier learned from hypothetical labeling in Section~\ref{sec::generalization-bound-kernel} where the discriminative similarity is induced by the error bound, and Section~\ref{sec::similarity-kdc} shows that a discriminative similarity can also be induced by kernel density classification. The application of the discriminative similarity learning framework to data clustering and semi-supervised learning is shown in Section~\ref{sec::application}, and we conclude the paper in Section~\ref{sec::conclusion}. Throughout this paper the term kernel standards for PSD kernel if no special notes are made.

\section{Discriminative Similarity Framework}\label{sec:discriminative-similarity-framework}

\subsection{Discriminative Similarity Framework for Clustering}\label{sec:discriminative-similarity-framework-clustering}
The discriminative clustering literature \citep{XuNLS04,GomesKP10} has demonstrated the potential of multi-class classification for the clustering problem. 
Inspired by the natural connection between clustering and classification, we proposes the framework of learning discriminative similarity for clustering by unsupervised classification which models the clustering problem as a multi-class classification problem: a classifier is learned from the training data built by a hypothetical labeling, which can be any possible cluster labeling. The optimal hypothetical labeling is supposed to be the one such that its associated classifier has the minimum generalization error bound. To study the generalization bound for the classifier learned from hypothetical labeling, the concept of classification model is needed. Given unlabeled data $\{\bx_l\}_{l=1}^n$, a classification model $M_{\cY}$ is constructed for any hypothetical labeling $\cY = \{y_l\}_{l=1}^n$ as below:
\begin{MyDefinition}\label{def:classification-model-clustering}
The classification model corresponding to the hypothetical labeling $\cY = \{y_i\}_{i=l+1}^n$ for clustering is defined as $M_{\cY}= (\cS, \cP, f)$. $\cS=\{x_i, y_i\}_{i=1}^n$ are the labeled data by the hypothetical labeling, and $\cS$ are assumed to be i.i.d. samples drawn from the joint distribution $P_{XY}$ over the data $X \in \R^d$ and its class label $Y \in \left\{ {1,2,...,c} \right\}$. $\cP$ is the family comprised of all such distributions, i.e. $\cP = \{P_{XY} \colon \cS \,\, \stackrel{i.i.d.}{\sim}  \,\, P_{XY}\}$. $f$ is a classifier learned using the training data $\cS$. The generalization error of the classification model $M_{\cY}$ is defined as the generalization error of the classifier $f$ in $M_{\cY}$.
\end{MyDefinition}
The optimal hypothetical labeling minimizes the generalization error bound for the classification model. With $f$ being different classifiers, different discriminative clustering models can be derived. When SVMs is used in the above discriminative model, unsupervised SVM \citep{XuNLS04} is obtained. When two nonparametric classifiers, i.e. the nearest neighbor classifier and the plug-in classifier, are used in this framework, the unsupervised classification method in \citep{YangLYWH-discriminative-similarity14} is recovered.

\subsection{Discriminative Similarity Framework for Semi-Supervised Learning}\label{sec:discriminative-similarity-framework-ssl}
Similar to the case of clustering, the discriminative similarity framework for semi-supervised learning also searches for the optimal hypothetical labeling such that the associated classifier has minimum generalization error bound.

Suppose the data $\{\bx_1,\ldots,\bx_l,\bx_{l+1},\ldots,\bx_n\} \subseteq \R^d$ are comprised of labeled and unlabeled set, the first $l$ points have labels $\bl_i \in \{1,\ldots,c\}$ for $ 1 \le i \le l$ and $c$ is the number of classes. In the following text, $y_i$ is the label of $x_i$ for $i \in \{1,\ldots,n\}$, and $y_i = \bl_i$ for $i \in \{1,\ldots,l\}$. Given any hypothetical labeling $\cY = \{y_i\}_{i=l+1}^n$ for the unlabeled data, the classification model for semi-supervised learning is defined as below:
\begin{MyDefinition}\label{def:classification-model-ssl}
The classification model corresponding to the hypothetical labeling $\cY = \{y_i\}_{i=l+1}^n$ for semi-supervised learning is defined as $M_{\cY}= (\cS, \cP, f)$. $\cS=\{x_i, y_i\}_{i=1}^n$ are the labeled data by the hypothetical labeling, and $\cS$ are assumed to be i.i.d. samples drawn from the joint distribution $P_{XY}$ over the data $X \in \R^d$ and its class label $Y \in \left\{ {1,2,...,c} \right\}$. $\cP$ is the family comprised of all such distributions, i.e. $\cP = \{P_{XY} \colon \cS \,\, \stackrel{i.i.d.}{\sim}  \,\, P_{XY}\}$. $f$ is a classifier learned using the training data $\cS$. The generalization error of the classification model $M_{\cY}$ is defined as the generalization error of the classifier $f$ in $M_{\cY}$.
\end{MyDefinition}
Our framework of learning discriminative similarity for semi-supervised learning model searches for the optimal hypothetical labeling that minimizes the generalization error bound for the classification model defined above. Different specific discriminative semi-supervised learning models can be derived with $f$ being different classifiers. For example, when SVMs is used, a model with the same optimization problem as the well known Semi-Supervised or Transductive SVMs \citep{Vapnik1998,Joachims1999,Chapelle2008} is obtained \footnote{Note that Transductive SVMs, despite its name, also learns a inductive rule}.

\subsection{Discriminative Similarity induced by Kernel Classifier}\label{sec:unsupervised-kernel-classifier}
We employ kernel classifier in the proposed discriminative similarity learning framework in Section~\ref{sec:discriminative-similarity-framework-clustering} and Section~\ref{sec:discriminative-similarity-framework-ssl}, so as to induce the discriminative similarity parameterized with learnable weights of the kernel classifier. The kernel classifier is designed based on similarity learning methods. Balcan et al. \citep{Balcan08} proposes a classification method using general similarity functions, and the classification rule measures the similarity of the test data to each class then assigns the test data to the class such that the weighed average of the similarity between the test data and the training data belonging to that class is maximized over all the classes. In \citep{CortesMR13}, kernel function is used as the similarity function, and the generalization error of a kernel classifier is derived using a properly defined kernel margin, wherein the classifier uses average instead of weighed average when computing point-to-class similarity. Inspired by these similarity learning methods, we propose hypothesis $h(\cdot,y)$ to measure the similarity of datum $\bx$ to class $y$ and it uses kernel as the similarity function:
\begin{small}\begin{align}\label{eq:hypothesis}
&h(\bx,y) = \sum\limits_{i \colon y_i = y} {\balpha_i}{ K_h(\bx-{\bx_i})}
\end{align}\end{small}
%
where $K_{h}(\bx) = \exp(-\frac{\|\bx\|_2^2}{2{h^2}})$ is the isotropic Gaussian kernel (with the constant that makes unit integral omitted) with bandwidth $h$, $\{\balpha_i\}_{i=1}^n$ are the nonnegative weights that sum up to $1$. Instead of the expectation-based similarity in \citep{Balcan08}, hypothesis $h$ is a the finite sample-based similarity between datum $\bx$ and class $y$, leading to a tractable optimization problem, as shown in the next section. The similarity-based kernel classifier $f$ predicts the label of the datum $\bx$ as the one for which the point-to-class similarity is maximized, i.e. $f(\bx)=\argmax_{y \in \{1,\ldots,c\}} h(\bx,y)$.

The generalization error bound for the kernel classifier is derived in the following section. \textit{Note that we only need to derive the error bound for the kernel classifiers in the discriminative similarity framework for clustering. The reason is that the error bound for clustering also applies to the case of semi-supervised learning, as a certain amount of given labels do not affect the derivation of the generalization bound.}


\section{Discriminative Similarity from Generalization Bound for Kernel Classifier}\label{sec::generalization-bound-kernel}
In this section, the generalization error bound for the classification model in Definition~\ref{def:classification-model-clustering} and Definition~\ref{def:classification-model-ssl} with the kernel classifier is derived as a sum of discriminative similarity between the data from different classes.

To analyze the generalization bound for the kernel classifier $f$, the following notations are introduced. Let $\balpha = [\balpha_1,\ldots,\balpha_n]^{\top}$ be the nonzero weights that sum up to $1$, $\balpha^{(y)}$ be a $n \times 1$ column vector representing the weights belonging to class $y$ such that $\balpha_i^{(y)}$ is $\balpha_i$ if $y = y_i$, and $0$ otherwise. The margin of the labeled sample $(\bx,y)$ is defined as $m_h(\bx,y) = h(\bx,y) - \argmax_{y' \neq y} h(\bx,y')$, the sample $(\bx,y)$ is classified correctly if $m_h(\bx,y) \ge 0$. We then derive the generalization error bound for $f$ using the Rademacher complexity of the function class comprised of all the possible margin functions $m_h$. The Rademacher complexity \citep{Bartlett2003,Koltchinskii01} of a function class is defined below:
\begin{MyDefinition}\label{def:RC}
Let $\{\sigma_i\}_{i=1}^n$ be $n$ i.i.d. random variables such that $\Pr[\sigma_i = 1] = \Pr[\sigma_i = -1] = \frac{1}{2}$. The Rademacher complexity of a function class $\cA$ is defined as
\begin{small}\begin{align}\label{eq:RC}
&\cfrakR(\cA) = {\E}_{\{\sigma_i\},\{\bx_i\}}\left[\sup_{h \in \cA} {|\frac{1}{n} \sum\limits_{i=1}^n {\sigma_i}{h(\bx_i)} | } \right]
\end{align}\end{small}
\end{MyDefinition}
Let $\bK$ be the gram matrix of the data by the kernel $K_h$ with $\bK_{ij} = K_h(\bx_i - \bx_j)$. Based on the generalization analysis by \citet{koltchinskii2002}, Theorem~\ref{theorem::kernel-error} presents the generalization error bound for the unsupervised kernel classifier $f$  using the empirical error and the Rademacher complexity of the function class $\cH$. Inspired by the regularization on multiclass kernel-based vector machines \citep{CrammerS01}, we propose the regularization term $\Omega(\balpha) = {\sum\limits_{y=1}^c  { {\balpha^{(y)}}^{\top} {\bK} {\balpha^{(y)}} } }$ which is required to be bounded by $B^2$ for some $B>0$. Denote by $\cH_y$ the space of all the hypothesis $h(\cdot,y)$ associated with label $y$, i.e.
\begin{small}\begin{align}\label{eq:cHy}
&\cH_y = \{(\bx,y) \to \sum\limits_{i \colon y_i = y} {\balpha_i}{K_{h}(\bx - {\bx_i})} \colon \balpha \ge \bzero, \bone^{\top} \balpha = 1, \Omega(\balpha) \le B^2 \}, 1 \le y \le c
\end{align}\end{small}
\noindent and define the hypothesis space $\cH = \{(\bx,y) \to m_h(\bx,y) \colon h(\bx,y) \in \cH_y \}$. Lemma~\ref{lemma::RC-bound} shows that the Rademacher complexity of the properly defined functional class $\cH$ is bounded by the regularization term $\Omega(\balpha)$ with a large probability:
\begin{MyLemma}\label{lemma::RC-bound}
Define $\Omega(\balpha) = {\sum\limits_{y=1}^c  { {\balpha^{(y)}}^{\top} {\bK} {\balpha^{(y)}} } }$. When $\Omega(\balpha) \le B^2$ where $B$ is a positive constant, with probability at least $1-\delta$ over the data $\{\bx_i\}_{i=1}^n$, the Rademacher complexity of the function space $\cH$ satisfies
\begin{small}\begin{align}\label{eq:RC-bound}
&\cfrakR(\cH) \le \frac{(2c-1){c}}{\sqrt n}B + {\sqrt 2}{Bc}(2c-1)\sqrt{\frac{\ln{\frac{2}{\delta}}}{2n}}
\end{align}\end{small}
\end{MyLemma}

With the bounded Rademacher complexity of the function class $\cH$, Theorem~\ref{theorem::kernel-error} presents the generalization error bound for the kernel classifier $f$ in the discriminative similarity learning framework.
\begin{MyTheorem}\label{theorem::kernel-error}
(\textit{Error of the Kernel Classifier})
Given the classification model $M_{\cY}= (\cS, P_{XY}, f)$ in Definition~\ref{def:classification-model-clustering} or Definition~\ref{def:classification-model-ssl}, if $\Omega(\balpha) \le B^2$, then with probability $1-\delta$ over the labeled data $\cS$ with respect to any distribution in $P_{XY}$, the generalization error of the kernel classifier $f$ satisfies
\begin{small}\begin{align}\label{eq:kernel-error-theorem}
&R(f) = \Pr\left[Y \neq f(X)\right] \le {\hat R_{n}(f)}  + \frac{8(2c-1){c}}{\gamma \sqrt n}B + \Big(\frac{{8{\sqrt 2}Bc}(2c-1)}{\gamma} + 1 \Big)\sqrt{\frac{\ln{\frac{4}{\delta}}}{2n}}
\end{align}\end{small}
where ${\hat R_{n}(f)} = \frac{1}{n}  \sum\limits_{i=1}^n \Phi \Big(\frac{h(\bx_i,y_i) - \sum\limits_{y \neq y_i}{h(\bx_i,y)}}{\gamma} \Big)$ is the empirical error of $f$ on the labeled data, $\gamma >0$ is a constant and $\Phi$ is defined as
\begin{small}\begin{align}\label{eq:Phi}
\Phi(x) = \left\{
     \begin{array}{cl}
       1   &x < 0\\
       1-x &0 \le x \le 1 \\
       0   &x > 1 \\
     \end{array}
\right.
\end{align}\end{small}
Moreover, if $\gamma \ge c-1$, the empirical error ${\hat R_{n}(f)}$ is
\begin{small}\begin{align}\label{eq:kernel-error-similarity}
&{\hat R_{n}(f)} =  1 - \frac{1}{n\gamma} \sum\limits_{i,j=1}^n {\frac{\balpha_i + \balpha_j}{2}} {K_{h}(\bx_i - {\bx_j})} + \frac{1}{n\gamma} \sum\limits_{1 \le i < j \le n}^n 2({\balpha_i + \balpha_j}){K_{h}(\bx_i-\bx_j)} {\1}_{y_i \neq y_j}
\end{align}\end{small}
\end{MyTheorem}
With the error bound in Theorem~\ref{theorem::kernel-error}, searching for the optimal hypothetical labeling $\cY$ amounts to solving the following optimization problem
\begin{small}\begin{align}\label{eq:obj-dlp-1}
&\min_{\balpha \in \Lambda,\cY = \{y_i\}_{i=1}^n}\sum\limits_{1 \le i < j \le n}^n 2({\balpha_i + \balpha_j}){K_{h}(\bx_i-\bx_j)} {\1}_{y_i \neq y_j} - \sum\limits_{i,j=1}^n {\frac{\balpha_i + \balpha_j}{2}} {K_{h}(\bx_i-\bx_j)} + {\lambda} \Omega(\balpha)
\end{align}\end{small}
Where $\Lambda = \{\balpha \colon \balpha \ge \bzero, \bone^{\top} \balpha = 1\}$ is the feasible set for the weights of the kernel classifier, and $\lambda > 0$ is a weighting parameter. Similar to the optimization problem of SVMs \citep{Vapnik1998,Joachims1999}, there is a balancing parameter $\lambda$ that balances between the empirical error and the regularization term. Substituting $\Omega(\balpha)$ into (\ref{eq:obj-dlp-1}),
\begin{small}\begin{align}\label{eq:obj-cds-discrete}
&\min_{\balpha \in \Lambda,\cY = \{y_i\}_{i=1}^n}
\sum\limits_{1 \le i < j \le n}^n S_{ij}^{\rm ker}  {\1}_{y_i \neq y_j} - \sum\limits_{i,j=1}^n {\frac{\balpha_i + \balpha_j}{2}} {K_{h}(\bx_i-\bx_j)} + {\lambda} \balpha^{\top} {\bK} \balpha
\end{align}\end{small}
\noindent where
\begin{small}\begin{align}\label{eq:kernel-similarity}
&S_{ij}^{\rm ker} = 2(\balpha_i + \balpha_j - {\lambda}{\balpha_i}{\balpha_j}){K_{h}(\bx_i-\bx_j)}, 1 \le i,j \le n
\end{align}\end{small}
\noindent $\lambda$ is tuned such that $S_{ij}^{\rm ker} \ge 0$, e.g. $\lambda \le 2$. The first term of the objective function (\ref{eq:obj-cds-discrete}) is $\sum\limits_{1 \le i < j \le n}^n S_{ij}^{\rm ker} {K_{h}(\bx_i-\bx_j)} {\1}_{y_i \neq y_j}$, which is the sum of similarity between the data from different classes with \textit {$S_{ij}^{\rm ker}$ being the discriminative similarity between $\bx_i$ and $\bx_j$ induced by the generalization error bound (\ref{eq:kernel-error-theorem}).}
In the next section, it is shown that the discriminative similarity with the same form as that induced by our generalization analysis, namely (\ref{eq:kernel-similarity}), can also be induced from the perspective of kernel density classification by kernel density estimators with nonuniform weights. It supports the theoretical justification for the induced discriminative similarity (\ref{eq:kernel-similarity}) in this section.

\section{A Kernel Density Classification Perspective}\label{sec::similarity-kdc}
The discriminative similarity can also be induced from kernel density classification with varying weights on the data, and binary classification is considered in this section. For any classification model $M_{\cY}= (\cS, P_{XY}, f)$ with hypothetical labeling $\cY$ and the labeled data $\cS = \{\bx_i,y_i\}_{i=1}^n$, suppose the joint distribution $P_{XY}$ over $\cX \times \{1,2\}$ has probabilistic density function $p(\bx,y)$. Let $P_X$ be the induced marginal distribution over the data with probabilistic density function $p(\bx)$. Robust kernel density estimation methods \citep{GirolamiH03,KimS08,MahapatruniG11,KimS12} suggest the following kernel density estimator where the kernel contributions of different data points are reflected by different nonnegative weights that sum up to $1$:
\begin{small}\begin{align}\label{eq:nonuniform-kde}
&\hat p(\bx) ={\tau_0} \sum\limits_{i=1}^n {\balpha_i}{K_{h}(\bx - {\bx_i})}, \bone^{\top} \balpha = 1, \balpha \ge 0
\end{align}\end{small}
where $\tau_0 = \frac{1}{({2{\pi}})^{d/2}{h^d}}$. Based on (\ref{eq:nonuniform-kde}), it is straightforward to obtain the following kernel density estimator of the density function $p(\bx,y)$:
\begin{small}\begin{align}\label{eq:nonuniform-kde-xy}
&\hat p(\bx,y) ={\tau_0} \sum\limits_{i:y_i = y} {\balpha_i}{K_{h}(\bx - {\bx_i})}
\end{align}\end{small}
Kernel density classifier is learnt from the labeled data $\cS$ and constructed by kernel density estimators (\ref{eq:nonuniform-kde-xy}). Kernel density classifier resembles the Bayes classifier, and it classifies the test data $\bx$ based on the conditional label distribution $P(Y|X=\bx)$, or equivalently, $\bx$ is assigned to class $1$ if $\hat p(\bx,1)- \hat p(\bx,2) \ge 0$, otherwise it is assigned to class $2$. Intuitively, it is preferred that the decision function $\hat r(\bx,\balpha) = \hat p(\bx,1)- \hat p(\bx,2)$ is close to the true Bayes decision function $r = p(\bx,1)- p(\bx,2)$. \citep{GirolamiH03,KimS08} propose to use Integrated Squared Error (ISE) as the metric to measure the distance between the kernel density estimators and their true counterparts, and the oracle inequality is obtained that relates the performance of the $L_2$ classifier in \citep{KimS08} to the best possible performance of kernel density classifier in the same category. ISE is adopted in our analysis of kernel density classification, and the ISE between the decision function $\hat r$ and the true Bayes decision function $r$ is defined as
\begin{small}\begin{align}\label{eq:ise-def}
&{\rm ISE} (\hat r,r) = \|\hat r - r\|_{\cL_2}^2 = \int_{\R^d} {(\hat r - r)^2} dx
\end{align}\end{small}

The upper bound for the ISE ${\rm ISE} (r,\hat r)$ also induces discriminative similarity between the data from different classes, which is presented in the following theorem.
\begin{MyTheorem}\label{theorem::ise}
Let $n_1 = \sum\limits_{i=1} \1_{y_i=1}$ and $n_2 = \sum\limits_{i=1} \1_{y_i=2}$. With probability at least $1-2{n_2}\exp\big(-2(n-1)\varepsilon^2\big)-2{n}\exp\big(-2n\varepsilon^2\big)$ over the labeled data $\cS$, the ISE between the decision function $\hat r(\bx,\balpha)$ and the true Bayes decision function $r(\bx)$ satisfies
\begin{small}\begin{align}\label{eq:ise-theorem}
& {\rm ISE} (\hat r,r) \le \frac{\tau_0}{n} {\hat {\rm ISE}} (\hat r,r) + {\tau_1} K(\balpha) + 2{\tau_0}\Big(\frac{1}{n-1}+\varepsilon\Big)
\end{align}\end{small}
\noindent where
\begin{small}\begin{align}\label{eq:hat-ise}
&{\hat {\rm ISE}}(\hat r,r) = 4 {\sum\limits_{1 \le i < j \le n}(\balpha_i+\balpha_j) K_{h} (\bx_i - \bx_j) {\1}_{y_i \neq y_j} } - {\sum\limits_{i,j = 1}^ n (\balpha_i+\balpha_j) K_{h} (\bx_i - \bx_j) }
\end{align}\end{small}
\begin{small}\begin{align}\label{eq:reg-ise}
&K(\balpha) = {\balpha}^{\top} (\bK_{{\sqrt 2}h}){\balpha}  -
4\sum\limits_{1 \le i < j \le n} {\balpha_i}{\balpha_j} K_{{\sqrt 2}h} (\bx_i - \bx_j) {\1}_{y_i \neq y_j}
\end{align}\end{small}
and $\bK_{{\sqrt 2}h}$ is the gram matrix evaluated on the data $\{\bx_i\}_{i=1}^n$ with the kernel $K_{{\sqrt 2}h}$.
\end{MyTheorem}
Let $\lambda_1 > 0$ be a weighting parameter, then the cost function ${\hat {\rm ISE}} + \lambda_1 K(\balpha)$, designed according to the empirical term ${\rm ISE} (\hat r,r)$ and the regularization term $K(\balpha)$ in the ISE error bound (\ref{eq:ise-theorem}), can be expressed as
\begin{scriptsize}\begin{align*}
&{\hat {\rm ISE}} + {\lambda_1} K(\balpha)
& \le {\sum\limits_{1 \le i < j \le n} 4(\balpha_i+\balpha_j - {\lambda_1}{\balpha_i}{\balpha_j} ) K_{h} (\bx_i - \bx_j) {\1}_{y_i \neq y_j} } - {\sum\limits_{i,j = 1}^ n (\balpha_i+\balpha_j) K_{h} (\bx_i - \bx_j) } + {\lambda_1}{\balpha}^{\top} \bK_{{\sqrt 2}h}{\balpha}
\end{align*}\end{scriptsize}
wherein the first term is comprised of sum of similarity between data from different classes with similarity $S_{ij}^{\rm ise} = 4(\balpha_i+\balpha_j - {\lambda_1}{\balpha_i}{\balpha_j} )K_{h} (\bx_i - \bx_j)$, and $S_{ij}^{\rm ise}$ is the discriminative similarity induced by the ISE bound for kernel density classification. Note that $S_{ij}^{\rm ise}$ has the same form as the discriminative similarity $S_{ij}^{\rm ker}$ (\ref{eq:kernel-similarity}) induced by the kernel classifier in our discriminative similarity learning framework, up to a scaling constant and the choice of the balancing parameter.

\section{Extension to General Similarity-Based Classifier}\label{sec::generalization-bound-similarity}
We now consider using a general symmetric and continuous function $S \colon \cX \times \cX \to [-1,1]$ as the similarity function in the classification model in Definition~\ref{def:classification-model-clustering} or Definition~\ref{def:classification-model-ssl} which is not necessarily a PSD kernel. The hypothesis $h(\cdot,y)$ becomes $h_S(\bx,y) = \sum\limits_{i \colon y_i = y} {\balpha_i}{ S(\bx, {\bx_i})}$.
In order to analyze the generalization property of the classification rule using the general similarity function, we first investigate the properties of general similarity function and its relationship to PSD kernels in terms of eigenvalues and eigenfunctions of the associated integral operator. The integral operator $({L_S}f)(\bx) = \int {S(\bx,\bt)f(\bt)} d \bt$ is well defined. It can be verified that $L_S$ is a compact operator since $S$ is continuous. According to the spectral theorem in operator theory, there exists an orthogonal basis $\{\phi_1,\phi_2,\ldots\}$ of $\cL^2$ which are the eigenfunctions of $L_S$, where $\cL^2$ is the space of measurable functions which are defined over $\cX$ and square Lebesgue integrable. $\phi_k$ is the eigenfunction of $L_S$ with eigenvalue $\lambda_k$ if ${L_S}{\phi_k} = {\lambda_k}{\phi_k}$. The following lemma shows that under certain assumption on the eigenvalues and eigenfunctions of $L_S$, a general symmetric and continuous similarity can be decomposed into two PSD kernels.

\begin{MyLemma}{\label{lemma::decompose-similarity}}
Suppose $S \colon \cX \times \cX \to [-1,1]$ is a symmetric continuous function, and $\{\lambda_k\}$ and $\{\phi_k\}$ are the eigenvalues and eigenfunctions of $L_S$ respectively. Suppose $\sum\limits_{k \ge 1} {\lambda_k}|\phi_k(\bx)|^2 < C$ for some constant $C > 0$. Then $S(\bx,\bt) = \sum\limits_{k \ge 1} {\lambda_k}{\phi_k(\bx)}{\phi_k(\bt)}$ for any $\bx,\bt \in \cX$, and it can be decomposed as the difference between two positive semi-definite kernels, namely $S(\bx,\bt) = S^{+}(\bx,\bt) - S^{-}(\bx,\bt)$ with
\begin{small}\begin{align} \label{eq:S-decompose}
&S^{+}(\bx,\bt) = \sum\limits_{k \colon \lambda_k \ge 0} \lambda_k \phi_k(\bx) \phi_k (\bt)  \quad
S^{-}(\bx,\bt) = \sum\limits_{k:\lambda_k < 0} |\lambda_k| \phi_k(\bx) \phi_k (\bt) 
\end{align}\end{small}
\end{MyLemma}
\begin{MyRemark}
If $S$ is a kernel, i.e. all of the eigenvalues of $L_S$ are nonnegative, then $\sum\limits_{k \ge 1} {\lambda_k}|\phi_k(\bx)|^2 $ is bounded by a constant \citep{Felipe07-learning-theory}. Also, when $S(\bx, \cdot)$ is square Lebesgue integrable, $\sum\limits_{k \ge 1} |{\lambda_k}\phi_k(\bx)|^2 < \infty$. Lemma~\ref{lemma::decompose-similarity} allows for general similarity function with negative eigenvalues if $\sum\limits_{k \ge 1} {\lambda_k}|\phi_k(\bx)|^2$ is bounded.
\end{MyRemark}

Resembling the case that kernel serves as the similarity function in Section~\ref{sec::generalization-bound-kernel}, we use the regularization term to bound the Rademacher complexity for the classification rule using general similarity function. Let $\Omega^{+}(\balpha) = {\sum\limits_{y=1}^c  { {\balpha^{(y)}}^{\top} {\bS^{+}} {\balpha^{(y)}} } }$ and $\Omega^{-}(\balpha) = {\sum\limits_{y=1}^c  { {\balpha^{(y)}}^{\top} {\bS^{-}} {\balpha^{(y)}} } }$ with $[\bS^{+}]_{ij} = S^{+}(\bx_i,\bx_j)$ and $[\bS^{-}]_{ij} = S^{-}(\bx_i,\bx_j)$. Similar to the analysis of kernel classifier, the space $\cH_y$ of all the hypothesis $h(\cdot,y)$ associated with label $y$ is defined as
\begin{small}\begin{align}\label{eq:cHy-S}
&\cH_{S,y} = \{(\bx,y) \to \sum\limits_{i \colon y_i = y} {\balpha_i}{S(\bx,{\bx_i})} \colon \balpha \ge \bzero, \bone^{\top}\balpha =1, \Omega^{+}(\balpha) \le {B^{+}}^2,\Omega^{-}(\balpha) \le {B^{-}}^2\}, 1 \le y \le c
\end{align}\end{small}
with positive number $B^+$ and $B^-$ which bounds $\Omega^{+}$ and $\Omega^{-}$ respectively. Let the margin function be $m_{h_S}(\bx,y) = h_S(\bx,y) - \argmax_{y' \neq y} h_S(\bx,y')$, the hypothesis space be $\cH_S = \{(\bx,y) \to m_{h_S}(\bx,y) \colon h(\bx,y) \in \cH_{S,y} \}$, and the general similarity-based classifier $f_S$ predicts the label of the datum $\bx$ by $f_S(\bx)=\argmax_{y \in \{1,\ldots,c\}} h_S(\bx,y)$. We then present the main results in this section, which show the bound for the Rademacher complexity of the hypothesis space, i.e. $\cfrakR(\cH_S)$, and the generalization error of unsupervised general similarity-based classifier $f_S$.

\begin{MyLemma}\label{lemma::RC-bound-S}
Suppose the assumptions in Lemma~\ref{lemma::decompose-similarity} hold. Define $\Omega^{+}(\balpha) = {\sum\limits_{y=1}^c  { {\balpha^{(y)}}^{\top} {\bS^{+}} {\balpha^{(y)}} } }$ and $\Omega^{-}(\balpha) = {\sum\limits_{y=1}^c  { {\balpha^{(y)}}^{\top} {\bS^{-}} {\balpha^{(y)}} } }$. When $\Omega^{+}(\balpha) \le {B^{+}}^2$,$\Omega^{-}(\balpha) \le {B^{-}}^2$ for positive constants $B^+$ and $B^-$, $\sup_{\bx \in \cX} |S^{+}(\bx,\bx)| \le R^2$, $\sup_{\bx \in \cX} |S^{-}(\bx,\bx)| \le R^2$ for some $R > 0$, then with probability at least $1-\delta$ over the data $\{\bx_i\}_{i=1}^n$, the Rademacher complexity of the class $\cH_S$ satisfies
\begin{small}\begin{align}\label{eq:RC-bound-S}
&\cfrakR(\cH_S) \le \frac{R(2c-1){c}(B^{+} + B^{-})}{\sqrt n} + 2c(2c-1)(B^{+}+B^{-})R^2\sqrt{\frac{\ln{\frac{2}{\delta}}}{2n}}
\end{align}\end{small}
\end{MyLemma}

\begin{MyTheorem}\label{theorem::similarity-error}
(\textit{Error of the General Similarity-Based Classifier})
Suppose the assumptions in Lemma~\ref{lemma::decompose-similarity} hold. Given the classification model $M_{\cY}= (\cS, P_{XY}, f_S)$ in Definition~\ref{def:classification-model-clustering} or Definition~\ref{def:classification-model-ssl}, if $\Omega^{+}(\balpha) \le {B^{+}}^2$, $ \Omega^{-}(\balpha) \le {B^{-}}^2$, then with probability $1-\delta$ over the labeled data $\cS$ with respect to any distribution in $P_{XY}$, under the assumptions of Lemma~\ref{lemma::decompose-similarity} and Lemma~\ref{lemma::RC-bound-S} on $S$, $S^+$ and $S^-$, the generalization error of the general classifier $f_S$ satisfies
\begin{small}\begin{align}\label{eq:similarity-error-theorem}
&R(f_S) = \Pr\left[Y \neq f_S(X)\right] \le {\hat R_{n}(f_S)} + \frac{8R(2c-1){c}(B^{+} + B^{-})}{\gamma \sqrt n} + (\frac{16c(2c-1)(B^{+}+B^{-})R^2}{\gamma} + 1)\sqrt{\frac{\log{\frac{4}{\delta}}}{2n}}
\end{align}\end{small}
where ${\hat R_{n}(f_S)} = \frac{1}{n}  \sum\limits_{i=1}^n \Phi \Big(\frac{h_S(\bx_i,y_i) - \sum\limits_{y \neq y_i}{h_S(\bx_i,y)}}{\gamma} \Big)$ is the empirical error of $f_S$ on the labeled data, $\gamma >0$ is a constant and $\Phi$ is defined  in (\ref{eq:Phi}).
Moreover, if $\gamma \ge c$, the empirical error ${\hat R_{n}(f_S)}$ is
\begin{small}\begin{align}\label{eq:similarity-error-similarity}
&{\hat R_{n}(f_S)} =  1 - \frac{1}{n\gamma} \sum\limits_{i,j=1}^n {\frac{\balpha_i + \balpha_j}{2}} {S(\bx_i, {\bx_j})} + \frac{1}{n\gamma} \sum\limits_{1 \le i < j \le n}^n 2({\balpha_i + \balpha_j}){S(\bx_i,\bx_j)} {\1}_{y_i \neq y_j}
\end{align}\end{small}
\end{MyTheorem}
\begin{MyRemark}
When $S$ is a kernel, e.g. $S = K_h$, it can be verified that $S^{-} \equiv 0$, $S = S^{+}$, $\Omega^{+}(\balpha) = \Omega(\balpha)$ so we can let $B^{+} = B$, $B^{-} = 0$, and it follows that the dominant term ${\hat R_{n}(f_S)} + \frac{8R(2c-1){c}(B^{+} + B^{-})}{\gamma \sqrt n}$ in the error bound (\ref{eq:similarity-error-theorem}) for general similarity function reduces to ${\hat R_{n}(f)}  + \frac{8(2c-1){c}B}{\gamma \sqrt n}$, the dominant term in the bound (\ref{eq:kernel-error-theorem}) for kernel classifier (with $R=1$).
\end{MyRemark}
\begin{MyRemark}
When the decomposition $S = S^{+} - S^{-}$ exists and $S^{+}$, $S^{-}$ are PSD kernels, $S$ is the kernel of some Reproducing Kernel Kre{\u{\i}}n Space (RKKS) \citep{Mary03-dissertation}. \citep{Ong04-non-positive-kernels,Ong16-krein-space} analyze the problem of learning SVM-style classifiers with indefinite kernels from the Kre{\u{\i}}n space. However, their work does not show when and how an indefinite and general similarity function can have PSD decomposition, as well as the generalization analysis for the similarity-based classifier using such general indefinite function as similarity measure. Our analysis deals with these problems in Lemma~\ref{lemma::decompose-similarity} and Theorem~\ref{theorem::similarity-error}.
\end{MyRemark}
Minimizing the above bound for general continuous similarity leads to the formulation that minimizes ${\hat R_{n}(f_S)} + \lambda\big(\Omega^{+}(\balpha) + \Omega^{-}(\balpha)\big)$, i.e.

\begin{small}\begin{align}\label{eq:obj-cds-similarity-discrete}
&\min_{\balpha,\cY \colon \balpha \ge \bzero, \bone^{\top}\balpha = 1,\cY = \{y_i\}_{i=1}^n}
\sum\limits_{1 \le i < j \le n}^n S_{ij}^{\rm sim}  {\1}_{y_i \neq y_j} - \sum\limits_{i,j=1}^n {\frac{\balpha_i + \balpha_j}{2}} {S(\bx_i,\bx_j)} + {\lambda} (\balpha^{\top}{\bS^{+}}\balpha + \balpha^{\top}{\bS^{-}}\balpha)
\end{align}\end{small}
\noindent where ${\lambda}>0$ is the weighting parameter for the regularization term $\Omega^{+}(\balpha) + \Omega^{-}(\balpha)$, and
\begin{small}\begin{align}\label{eq:general-similarity}
&S_{ij}^{\rm sim} = 2(\balpha_i + \balpha_j ){S(\bx_i, \bx_j)} - 2{\lambda}{\balpha_i}{\balpha_j}S^{+}(\bx_i, \bx_j) - 2{\lambda}{\balpha_i}{\balpha_j}S^{-}(\bx_i, \bx_j), 1 \le i,j \le n
\end{align}\end{small}
is the discriminative similarity between data from different classes, which is induced by the generalization error bound for the general similarity-based classifier $f_S$. When $S$ is a kernel, $S^{-} \equiv 0$, $S = S^{+}$, then $S_{ij}^{\rm sim}$ reduces to $S_{ij}^{\rm ker}$ in (\ref{eq:kernel-similarity}), the similarity induced by the kernel classifier.

\begin{MyRemark}[Similarity Machines: SVM-Type Classifier with General Similarity Function]
Traditional Kernel Support Vector Machines (Kernel SVMs), one of the most representative of Kernel Machine methods, maps the data into a infinite dimensional Reproducing Kernel Hilbert Space (RKHS) associated with the chosen kernel, and learn max-margin linear classifier in RKHS. The mapping function is a Mercer Kernel in most cases, which is symmetric, continuous and positive semi-definite. The past several decades have witnessed the great success of SVMs in solid theoretical foundation of statistical learning, and broad applications in a vast regime of machine learning, pattern recognition. However, the requirement of positive semi-definiteness for Mercer Kernel substantially restricts the feasibility of SVMs for learning max-margin classifier with general similarity function which is not necessarily a Mercer kernel. Based on Lemma~\ref{lemma::decompose-similarity}, we can propose Similarity Machines as a generalization of Kernel SVMs, which is a framework of learning maximum margin classifier with general similarity function $S$ which is symmetric but not necessarily positive semi-definite. Similarity Machines has generalization error bound which reduces to the canonical error bound for Kernel SVMs when the similarity function is in fact a PSD kernel. The parameters of Similarity Machines can be obtained by minimizing the objective function based on its error bound. More details about the generalization analysis of Similarity Machines are included in the appendix of this paper.
\end{MyRemark}

\section{Applications}
\label{sec::application}
In this section, we present new clustering and semi-supervised learning method using the discriminative similarity induced by the kernel classifier.
\subsection{Application to Data Clustering}\label{sec::application-clustering}
We propose a novel data clustering method named Clustering by Discriminative Similarity via Kernel classification (CDSK) which is based on our discriminative similarity learning framework with kernel classifier. CDSK aims to minimize (\ref{eq:obj-cds-discrete}). However, problem (\ref{eq:obj-cds-discrete}) involves minimization with respect to discrete cluster labels $\cY = \{y_i\}$ which is NP-hard. In addition, it potentially results in a trivial solution which puts all the data in a single cluster due to the lack of constraints on the cluster balance. Therefore, (\ref{eq:obj-cds-discrete}) is relaxed in the proposed optimization problem for CDSK below:
\begin{small}\begin{align}\label{eq:obj-cdsk}
&\min_{\balpha \in \Lambda, \bY \in \R^{n \times c }}
{\rm Tr} (\bY^{\top} \bL^{\rm ker} \bY) - \sum\limits_{i,j=1}^n {\frac{\balpha_i + \balpha_j}{2}} {K_{h}(\bx_i-\bx_j)} + {\lambda} \balpha^{\top} {\bK} \balpha
 \quad s.t. \,\, \bY^{\top} \bD^{\rm ker} \bY  = \bI_n
\end{align}\end{small}
where $\Lambda = \{\balpha \colon \balpha \ge \bzero, \bone^{\top} \balpha = 1\}$, $[\bS^{\rm ker}]_{ij} = S_{ij}^{\rm ker}$, $\bL^{\rm ker} = \bD^{\rm ker} - \bS^{\rm ker}$ is the graph Laplacian computed with $\bS^{\rm ker}$, $\bD^{\rm ker}$ is a diagonal matrix with each diagonal element being the sum of the corresponding row of $\bS^{\rm ker}$: $[\bD^{\rm ker}]_{ii} = \sum\limits_{j=1}^n \bS_{ij}^{\rm ker}$, $\bI_n$ is a $n \times n$ identity matrix. Note that when each column of $\bY$ is a binary membership indicator vector for the corresponding cluster, ${\rm Tr}(\bY^{\top} \bL^{\rm ker} \bY) = \sum\limits_{1 \le i < j \le n}^n S_{ij}  {\1}_{y_i \neq y_j}$. Similar to spectral clustering \citep{Ng01}, the constraint $\bY^{\top} \bD^{\rm ker} \bY  = \bI_n$ prevents imbalanced data clusters.

Problem (\ref{eq:obj-cdsk}) is optimized by coordinate descent. In each iteration of coordinate descent, optimization with respect to $\bY$ is performed with fixed $\balpha$, which is exactly the same problem as that of spectral clustering with a solution formed by the smallest $c$ eigenvectors of the normalized graph Laplacian $(\bD^{\rm ker})^{-1/2}{\bL^{\rm ker}}(\bD^{\rm ker})^{-1/2}$; then the optimization with respect to $\balpha$ is performed with fixed $\bY$, which is a standard constrained quadratic programming problem. The iteration of coordinate descent proceeds until convergence or the maximum iteration number is achieved.

\subsection{Application to Semi-Supervised Learning}\label{sec::application-clustering}
We also propose a new semi-supervised learning method based on Label Propagation using the Discriminative Similarity induced by the Kernel classification (LPDSK). The formulation of LPDSK is 
\begin{small}\begin{align}\label{eq:obj-lpdsk}
&\min_{\balpha \in \Lambda, \bY \in \R^{n \times c }}
{\rm Tr} (\bY^{\top} \bL^{\rm ker} \bY) - \sum\limits_{i,j=1}^n {\frac{\balpha_i + \balpha_j}{2}} {K_{h}(\bx_i-\bx_j)} + {\lambda} \balpha^{\top} {\bK} \balpha
 \quad &s.t. \,\, \bY_{ij} = \bF_{ij} \,\, {\rm for } \,\, 1 \le i \le l
\end{align}\end{small}%
where $\bF$ is a matrix of $n \times c$ with its elements set by the given labels, i.e. $\bF_{ij} = 1$ if $\bx_i$ has label $y_i = j$  for $1 \le i \le l$, otherwise $\bF_{ij} = 0$.

Similar to the case of clustering, Problem (\ref{eq:obj-lpdsk}) is also optimized by coordinate descent. In each iteration of coordinate descent, optimization with respect to $\bY$ is performed with fixed $\balpha$. With the block representation $\bF = \left[ {\begin{array}{*{20}{c}}
\bF_l\\
\bF_u
\end{array}} \right]$, $\bY = \left[ {\begin{array}{*{20}{c}}
\bY_l\\
\bY_u
\end{array}} \right]$, $\bL = \left[ {\begin{array}{*{20}{c}}
\bL_{ll}^{\rm ker}&\bL_{lu}^{\rm ker}\\
\bL_{ul}^{\rm ker}&\bL_{uu}^{\rm ker}
\end{array}} \right]$ where $\bF_l$ and $\bY_l$ are of size $l \times c$ , $\bL_{ll}^{\rm ker}$ is of size $l \times l$ and $\bL_{lu}^{\rm ker}$ is of size $l \times (n-l)$, it can be verified that this subproblem admits a closed form solution $\bY_u = - (\bL^{\rm ker}_{uu})^{-1} \bL^{\rm ker}_{ul} \bF_l$ when $\bL^{\rm ker}_{uu}$ is invertible; the optimization with respect to $\balpha$ is the same as the case of clustering for problem (\ref{eq:obj-cdsk}). The iteration of coordinate descent proceeds until convergence or the maximum iteration number is achieved.

\section{Conclusions}\label{sec::conclusion}
We propose a novel discriminative similarity learning framework wherein the discriminative similarity is induced by the generalization error bound for the classifier learned from hypothetical labeling, and the optimal hypothetical labeling is pursued by minimizing the generalization bound for the associated classifier. A kernel classifier is employed in the proposed framework. The learnable weights in discriminative similarity induced by the kernel classifier allows for adaptive similarity accommodating the local variation of the data. We also analyze the generalization property of similarity-based classifier with general continuous similarity function rather than a PSD kernel. We present new clustering and semi-supervised learning method based on the discriminative similarity learning framework with the kernel classifier, i.e. clustering by discriminative similarity via kernel classification (CDSK) and label propagation by discriminative similarity via kernel classification (LPDSK).


\small{
\bibliographystyle{unsrt}
\bibliography{mybib}
}

\section{Appendix}
\label{sec::appendix}
\subsection{Proof of Lemma~\ref{lemma::RC-bound}}
\begin{proof}
Inspired by \cite{koltchinskii2002}, we first prove that the Rademacher complexity of the function class formed by the maximum of several hypotheses is bounded by two times the sum of the Rademacher complexity of the function classes that these hypothesis belong to, i.e.
\begin{small}\begin{align}\label{eq:max-RC-bound}
\cfrakR(\cH_{\max}) \le 2\sum\limits_{y=1}^k \cfrakR(\cH_y)
\end{align}\end{small}
\noindent where $\cH_{\max} = \{\max\{h_1,\ldots,h_k\} \colon h_y \in \cH_y,  1 \le y \le k \}$ for $1 \le k \le c-1$.
If no confusion arises, the notations $(\{\sigma_i\},\{\bx_i,y_i\})$ are omitted in the subscript of the expectation operator in the following text, i.e. ${\E}_{\{\sigma_i\},\{\bx_i,y_i\}}$ is abbreviated to ${\E}$.
According to Theorem $11$ of \cite{koltchinskii2002}, it can be verified that
\begin{small}\begin{align*}
&{\E}_{\{\sigma_i\},\{\bx_i,y_i\}}\Big[\big(\sup_{h \in \cH_{\max}} {|\frac{1}{n} \sum\limits_{i=1}^n {\sigma_i}{h(\bx_i)} | }\big)^{+} \Big]
\le \sum\limits_{y=1}^k {\E}_{\{\sigma_i\},\{\bx_i,y_i\}}\Big[\big(\sup_{h \in \cH_y} {|\frac{1}{n} \sum\limits_{i=1}^n {\sigma_i}{h(\bx_i)} | }\big)^{+} \Big]
\end{align*}\end{small}
Therefore,
{\allowdisplaybreaks
\begin{small}\begin{align}\label{eq:kernel-error-seg1}
&\cfrakR(\cH_{\max}) = {\E}_{\{\sigma_i\},\{\bx_i,y_i\}}\Big[\sup_{h \in \cH_{\max}} {|\frac{1}{n} \sum\limits_{i=1}^n {\sigma_i}{h(\bx_i)} | } \Big] \nonumber \\
& \le {\E}_{\{\sigma_i\},\{\bx_i,y_i\}}\Big[\big(\sup_{h \in \cH_{\max}} {\frac{1}{n} \sum\limits_{i=1}^n {\sigma_i}{h(\bx_i)} }\big)^{+} \Big] +
{\E}_{\{\sigma_i\},\{\bx_i,y_i\}}\Big[\big(\sup_{h \in \cH_{\max}} {-\frac{1}{n} \sum\limits_{i=1}^n {\sigma_i}{h(\bx_i)} }\big)^{+} \Big] \nonumber \\
&=2 {\E}_{\{\sigma_i\},\{\bx_i,y_i\}}\Big[\big(\sup_{h \in \cH_{\max}} {\frac{1}{n} \sum\limits_{i=1}^n {\sigma_i}{h(\bx_i)} }\big)^{+} \Big] \nonumber \\
&\le 2 \sum\limits_{y=1}^k {\E}_{\{\sigma_i\},\{\bx_i,y_i\}}\Big[\big(\sup_{h \in \cH_y} {\frac{1}{n} \sum\limits_{i=1}^n {\sigma_i}{h(\bx_i)} }\big)^{+} \Big] \nonumber \\
&\le 2 \sum\limits_{y=1}^k {\E}_{\{\sigma_i\},\{\bx_i,y_i\}}\Big[\sup_{h \in \cH_y} {|\frac{1}{n} \sum\limits_{i=1}^n {\sigma_i}{h(\bx_i)}| } \Big] = 2 \sum\limits_{y=1}^k \cfrakR(\cH_y)
\end{align}\end{small}
}
And the equality in the third line of (\ref{eq:kernel-error-seg1}) is due to the fact that $-\sigma_i$ has the same distribution as $\sigma_i$. Using this fact again, (\ref{eq:max-RC-bound}), we have
{\allowdisplaybreaks
\begin{small}\begin{align}\label{eq:kernel-error-seg2}
&\cfrakR(\cH) = {\E}_{\{\sigma_i\},\{\bx_i,y_i\}}\Big[\sup_{m_h \in \cH} {\big|\frac{1}{n} \sum\limits_{i=1}^n {\sigma_i}{m_h(\bx_i,y_i)} \big| } \Big] \nonumber \\
&= {\E}_{\{\sigma_i\},\{\bx_i,y_i\}}\Big[\sup_{m_h \in \cH} {\big|\frac{1}{n} \sum\limits_{i=1}^n {\sigma_i} \sum\limits_{y=1}^c  {m_h(\bx_i,y)}{{\1}_{y=y_i}}\big| } \Big] \nonumber \\
& \le \sum\limits_{y=1}^c {\E}_{\{\sigma_i\},\{\bx_i,y_i\}}\Big[\sup_{m_h \in \cH} {\big|\frac{1}{n} \sum\limits_{i=1}^n {\sigma_i} {m_h(\bx_i,y)}{{\1}_{y=y_i}} \big| } \Big] \nonumber \\
&\le \frac{1}{2n} \sum\limits_{y=1}^c {\E}_{\{\sigma_i\},\{\bx_i,y_i\}}\Big[\sup_{m_h \in \cH} {\big | \sum\limits_{i=1}^n {\sigma_i} {m_h(\bx_i,y)}(2{\1}_{y=y_i}-1) \big| } \Big] + \frac{1}{2n} \sum\limits_{y=1}^c {\E}_{\{\sigma_i\},\{\bx_i\}}\Big[\sup_{m_h \in \cH} {\big| \sum\limits_{i=1}^n {\sigma_i} {m_h(\bx_i,y)}\big| } \Big] \nonumber \\
&= \frac{1}{n} \sum\limits_{y=1}^c {\E_{\{\sigma_i\},\{\bx_i\}}} \Big[\sup_{m_h \in \cH} {\big| \sum\limits_{i=1}^n {\sigma_i} {m_h(\bx_i,y)}\big| } \Big]
\end{align}\end{small}
}
Also, for any given $1 \le y \le c$
{\allowdisplaybreaks
\begin{small}\begin{align}\label{eq:kernel-error-seg3}
&\frac{1}{n} {\E_{\{\sigma_i\},\{\bx_i\}}}\Big[\sup_{m_h \in \cH} {| \sum\limits_{i=1}^n {\sigma_i} {m_h(\bx_i,y)}| } \Big] \nonumber \\
&= \frac{1}{n} {\E_{\{\sigma_i\},\{\bx_i\}}}\Big[\sup_{h(\cdot,y) \in \cH_y, y=1 \ldots c} {| \sum\limits_{i=1}^n {\sigma_i} {h(\bx_i,y) - {\sigma_i} \argmax_{y' \neq y} h(\bx_i,y') }| } \Big] \nonumber \\
& \le \frac{1}{n} {\E_{\{\sigma_i\},\{\bx_i\}}}\Big[\sup_{h(\cdot,y) \in \cH_y} {| \sum\limits_{i=1}^n {\sigma_i} {h(\bx_i,y)}| } \Big] +
\frac{1}{n} {\E_{\{\sigma_i\},\{\bx_i\}}}\Big[\sup_{h(\cdot,y') \in \cH_y', y' \neq y } {| \sum\limits_{i=1}^n {\sigma_i} {\argmax_{y' \neq y} h(\bx_i,y')} | } \Big] \nonumber \\
& \le \frac{1}{n} {\E_{\{\sigma_i\},\{\bx_i\}}}\Big[\sup_{h(\cdot,y) \in \cH_y} {| \sum\limits_{i=1}^n {\sigma_i} {h(\bx_i,y)}| } \Big] +
\frac{2}{n} \sum\limits_{y' \neq y} {\E_{\{\sigma_i\},\{\bx_i\}}}\Big[\sup_{h(\cdot,y') \in \cH_y'} {| \sum\limits_{i=1}^n {\sigma_i} {h(\bx_i,y')} | } \Big]
\end{align}\end{small}
}
Combining (\ref{eq:kernel-error-seg2}) and (\ref{eq:kernel-error-seg3}),
{\allowdisplaybreaks
\begin{small}\begin{align}\label{eq:kernel-error-seg4}
&\cfrakR(\cH) \le \sum\limits_{y=1}^c \Bigg(\frac{1}{n} {\E_{\{\sigma_i\},\{\bx_i\}}}\Big[\sup_{h(\cdot,y) \in \cH_y} {| \sum\limits_{i=1}^n {\sigma_i} {h(\bx_i,y)}| } \Big] +
\frac{2}{n} \sum\limits_{y' \neq y} {\E_{\{\sigma_i\},\{\bx_i\}}}\Big[\sup_{h(\cdot,y') \in \cH_y'} {| \sum\limits_{i=1}^n {\sigma_i} {h(\bx_i,y')} | } \Big] \Bigg) \nonumber \\
& = (2c-1)\sum\limits_{y=1}^c {\E_{\{\sigma_i\},\{\bx_i\}}}\Big[\sup_{h(\cdot,y) \in \cH_y} {| \frac{1}{n}\sum\limits_{i=1}^n {\sigma_i} {h(\bx_i,y)}| } \Big]
= (2c-1)\sum\limits_{y=1}^c \cfrakR(\cH_y)
\end{align}\end{small}
}
Therefore, the Rademacher complexity of $\cH$ is upper bounded by $(2c-1)$ times the sum of the Rademacher complexity of the hypothesis classes $\{\cH_y\}_{y=1}^c$. Next, we derive the upper bound for the sum of the Rademacher complexity of all the hypothesis classes, namely $\sum\limits_{y=1}^c \cfrakR(\cH_y)$. Denote by $\phi$ the feature mapping for the kernel $K_h$ which takes value in the Reproducing Kernel Hilbert Space $H_K$ associated with $K_h$ and satisfies $\langle\phi(\bx), \phi(\bt)\rangle_{H_K} = K_h(\bx-\bt)$ for $\bx, \bt \in \R^d$.

For
$h(\cdot,y) \in \cH_y$
\begin{small}\begin{align*}
&h(\bx,y) = \sum\limits_{i \colon y_i = y} {\balpha_i}{ K_h(\bx-{\bx_i})} = \langle \bw, \phi(\bx) \rangle
\end{align*}\end{small}
with $\bw = \sum\limits_{i \colon y_i = y} {\balpha_i} \phi(\bx_i)$, and $\|\bw\|_{H_K}^2 = {\balpha^{(y)}}^{\top} {\bK} {\balpha^{(y)}} \le B^2 \Rightarrow \|\bw\|_{H_K} \le B$. It follows that
\begin{small}\begin{align*}
\cH_y \subseteq \tilde \cH_y \triangleq \{(\bx,y) \to \langle \bw, \phi(\bx) \rangle_{H_K} \colon \|\bw\|_{H_K} \le B \}, 1 \le y \le c
\end{align*}\end{small}
And it follows that $\cfrakR(\cH_y) \le \cfrakR(\tilde \cH_y)$. Because we are deriving upper bound for $\cfrakR(\cH_y)$, in the following text of the proof we slightly abuse the notation and let $\cH_y$ represents $\tilde \cH_y$ if no confusion arises. Note that  for any $h \in \cH_y$, $h(\bx) - h(\bt) = \langle \bw, \phi(\bx) - \phi(\bt) \rangle_{H_K} \le \|\bw\|_{H_K} \|\phi(\bx) - \phi(\bt)\|_{H_K} \le \sqrt{2}B$.
We also have
\begin{small}\begin{align}\label{eq:kernel-error-seg5}
& \sum\limits_{y=1}^c \cfrakR(\cH_y) = \sum\limits_{y=1}^c {\E_{\{\sigma_i\},\{\bx_i\}}}\bigg[\sup_{h(\cdot,y) \in \cH_y} {\Big| \frac{1}{n}\sum\limits_{i=1}^n {\sigma_i} {h(\bx_i,y)} \Big| } \bigg]
\end{align}\end{small}
Similar to Theorem $11$ in \cite{Bartlett2003}, we now approximate the Rademacher complexity of the function class $\cH_y$, i.e. $\cfrakR(\cH_y)$, with its empirical version $\hat \cfrakR(\cH_y)$ using the given sample $\{\bx_i\}$. For each $1 \le y \le c$, Define $E_{\{\bx_i\}}^{(y)} = \hat \cfrakR(\cH_y) = {\E_{\{\sigma_i\}}}\bigg[\sup_{h(\cdot,y) \in \cH_y} {\Big| \frac{1}{n}\sum\limits_{i=1}^n {\sigma_i} {h(\bx_i,y)} \Big| } \bigg]$, then $\sum\limits_{y=1}^c \cfrakR(\cH_y) = \E_{\{\bx_i\}} \Big[\sum\limits_{y=1}^c E_{\{\bx_i\}}^{(y)}\Big]$, and
\begin{small}\begin{align*}
&\sup_{\bx_1,\ldots, \bx_n, \bx_t^{'}} \Big| E_{\bx_1,\ldots,\bx_{t-1},\bx_t,\bx_{t+1},\ldots,\bx_n}^{(y)} - E_{\bx_1,\ldots,\bx_{t-1},\bx_t^{'},\bx_{t+1},\ldots,\bx_n}^{(y)}\Big| \\
& = \sup_{\bx_1,\ldots, \bx_n, \bx_t^{'}}\Bigg| {\E_{\{\sigma_i\}}}\bigg[\sup_{h(\cdot,y) \in \cH_y} {\Big| \frac{1}{n}\sum\limits_{i=1}^n {\sigma_i} {h(\bx_i,y)} \Big| }
-\sup_{h(\cdot,y) \in \cH_y} \Big| \frac{1}{n}\sum\limits_{i \neq t} {\sigma_i} {h(\bx_i,y)} + \frac{h(\bx_t^{'},y)}{n} \Big|
\bigg] \Bigg| \\
&\le \sup_{\bx_1,\ldots, \bx_n, \bx_t^{'}}{\E_{\{\sigma_i\}}}\Bigg[ \bigg| \sup_{h(\cdot,y) \in \cH_y} \Big|{\frac{1}{n}\sum\limits_{i=1}^n {\sigma_i} {h(\bx_i,y)} \Big| }
-\sup_{h(\cdot,y) \in \cH_y} \Big| \frac{1}{n}\sum\limits_{i \neq t} {\sigma_i} {h(\bx_i,y)} + \frac{h(\bx_t^{'},y)}{n} \Big| \bigg|
\Bigg]  \\
&\le \sup_{\bx_1,\ldots, \bx_n, \bx_t^{'}}{\E_{\{\sigma_i\}}}\Bigg[ \sup_{h(\cdot,y) \in \cH_y} \bigg|  \Big|{\frac{1}{n}\sum\limits_{i=1}^n {\sigma_i} {h(\bx_i,y)} \Big| }
-\Big| \frac{1}{n}\sum\limits_{i \neq t} {\sigma_i} {h(\bx_i,y)} + \frac{h(\bx_t^{'},y)}{n} \Big| \bigg|
\Bigg] \\
&\le \sup_{\bx_1,\ldots, \bx_n, \bx_t^{'}}{\E_{\{\sigma_i\}}}\Bigg[ \sup_{h(\cdot,y) \in \cH_y} \bigg|  \frac{1}{n}\sum\limits_{i=1}^n {\sigma_i} {h(\bx_i,y)}
-\Big( \frac{1}{n}\sum\limits_{i \neq t} {\sigma_i} {h(\bx_i,y)} + \frac{h(\bx_t^{'},y)}{n} \Big) \bigg|
\Bigg] \\
&= \sup_{\bx_t, \bx_t^{'}} {\E_{\{\sigma_i\}}}\Bigg[ \sup_{h(\cdot,y) \in \cH_y} \bigg|  \frac{h(\bx_t,y)}{n} - \frac{h(\bx_t^{'},y)}{n}
\bigg|
\Bigg] \le \frac{\sqrt{2}B}{n}
\end{align*}\end{small}
And it follows that $\Big| \sum\limits_{y=1}^c E_{\bx_1,\ldots,\bx_{t-1},\bx_t,\bx_{t+1},\ldots,\bx_n}^{(y)} - \sum\limits_{y=1}^c E_{\bx_1,\ldots,\bx_{t-1},\bx_t^{'},\bx_{t+1},\ldots,\bx_n}^{(y)}\Big| \le \frac{\sqrt{2}Bc}{n}$. According to the McDiarmid's Inequality,
\begin{small}\begin{align}\label{eq:RC-empirical}
\Pr\Big[ \big| \sum\limits_{y=1}^c \hat \cfrakR(\cH_y) - \sum\limits_{y=1}^c \cfrakR(\cH_y)  \big| \ge \varepsilon \Big] \le 2\exp\big(-\frac{n\varepsilon^2}{B^2c^2}\big)
\end{align}\end{small}
Now we derive the upper bound for the empirical Rademacher complexity:
{\allowdisplaybreaks
\begin{small}\begin{align}\label{eq:kernel-error-seg5}
&\sum\limits_{y=1}^c \hat \cfrakR(\cH_y) = \sum\limits_{y=1}^c {\E_{\{\sigma_i\}}}\bigg[\sup_{ \|\bw\|_{H_K} \le B } {\Big| \frac{1}{n}\sum\limits_{i=1}^n {\sigma_i} \langle \bw, \phi(\bx_i) \rangle \Big| } \bigg] \nonumber \\
&=\sum\limits_{y=1}^c \sum\limits_{y=1}^c {\E_{\{\sigma_i\}}}\bigg[\sup_{ \|\bw\|_{H_K} \le B } {\Big| \frac{1}{n} \langle \bw, \sum\limits_{i=1}^n {\sigma_i} \phi(\bx_i) \rangle \Big| } \bigg] \nonumber \\
&\le \frac{B}{n}  \sum\limits_{y=1}^c {\E_{\{\sigma_i\}}}\bigg[{ \|\sum\limits_{i=1}^n {\sigma_i} \phi(\bx_i) \|_{H_K}  } \bigg] \nonumber \\
&\le \frac{B}{n}  \sum\limits_{y=1}^c \Big({\E_{\{\sigma_i\}}}\bigg[{\|\sum\limits_{i=1}^n {\sigma_i} \phi(\bx_i) \|^2_{H_K}  } \bigg] \Big)^{\frac{1}{2}} \nonumber \\
& = \frac{B}{n}  \sum\limits_{y=1}^c \Big( \sum\limits_{i=1}^n K_h(\bx_i-\bx_i) \Big)^{\frac{1}{2}} \nonumber \\
& = \frac{Bc}{\sqrt{n}}
\end{align}\end{small}
}
\noindent where $\bK$ is a $n \times n$ gram matrix with $\bK_{ij} = K_h(\bx_i - \bx_j)$, $\balpha^{(y)}$ is a $n \times 1$ column vector such that $\balpha_i^{(y)}$ is $\balpha_i$ if $y_i = y$, and $0$ otherwise. By (\ref{eq:kernel-error-seg4}), (\ref{eq:RC-empirical}) and (\ref{eq:kernel-error-seg5}), with probability at least $1-\delta$, we have
\begin{small}\begin{align}\label{eq:kernel-error-seg6}
&\cfrakR(\cH) \le (2c-1)\sum\limits_{y=1}^c \cfrakR(\cH_y) \nonumber \\
&\le \frac{(2c-1){c}B}{\sqrt n} + {\sqrt 2}{Bc}(2c-1)\sqrt{\frac{\ln{\frac{2}{\delta}}}{2n}}
\end{align}\end{small}
\end{proof}

\subsection{Proof of Theorem~\ref{theorem::kernel-error}}
\begin{proof}
According to Theorem 2 in \cite{koltchinskii2002}, with probability $1-\delta$ over the labeled data $\cS$ with respect to any distribution in $\cP$, the generalization error of the kernel classifier $f$ satisfies
\begin{small}\begin{align}\label{eq:kernel-error-lemma}
&{R(f)} \le {\hat R_{n}(f)}+ \frac{8}{\gamma}\cfrakR(\cH) + \sqrt{\frac{\ln{2/\delta}}{2n}}
\end{align}\end{small}
\noindent where ${\hat R_{n}(f)} = \frac{1}{n} \sum\limits_{i=1}^n {\Phi}(\frac{m_h(\bx_i,y_i)}{\gamma})$ is empirical error of the classifier for $\gamma > 0$. Note that ${\Phi}(\frac{m_h(\bx_i,y_i)}{\gamma}) \le \Phi \Big(\frac{h(\bx_i,y_i) - \sum\limits_{y \neq y_i}{h(\bx_i,y)}}{\gamma} \Big)$, applying Lemma~\ref{lemma::RC-bound}, (\ref{eq:kernel-error-theorem}) holds with probability $1-\delta$. When $\gamma \ge c-1$, it can be verified that $\Big|h(\bx_i,y_i) - \sum\limits_{y \neq y_i}{h(\bx_i,y)}\Big| \le c-1$ for all $(\bx_i,y_i)$, so that $$\Phi \Big(\frac{h(\bx_i,y_i) - \sum\limits_{y \neq y_i}{h(\bx_i,y)}}{\gamma} \Big) = 1-\frac{h(\bx_i,y_i) - \sum\limits_{y \neq y_i}{h(\bx_i,y)}}{\gamma}$$ and (\ref{eq:kernel-error-similarity}) is obtained.
\end{proof}

\subsection{Proof of Theorem~\ref{theorem::ise}}
\begin{proof}
According to definition of ISE,
\begin{small}\begin{align}\label{eq:ise-theorem-seg1}
&{\rm ISE} (\hat r,r) = \int_{\R^d} {(\hat r - r)^2} dx = \int_{\R^d} {{\hat r(\bx,\balpha)}^2} dx - 2\int_{\R^d} {\hat r(\bx,\balpha)} r(\bx) dx + \int_{\R^d} {r(\bx)}^2 dx
\end{align}\end{small}
For a given distribution, $\int_{\R^d} {r(\bx)}^2 dx$ is a constant. By Gaussian convolution theorem,
\begin{small}\begin{align}\label{eq:ise-theorem-seg2}
&\int_{\R^d} {{\hat r(\bx,\balpha)}^2} dx = {\tau_1} \sum\limits_{y=1}^2 {\balpha^{(y)}}^{\top} (\bK_{{\sqrt 2}h}){\balpha^{(y)}} -
{\tau_1} \sum\limits_{1 \le i < j \le n} 2{\balpha_i}{\balpha_j} {\bK}_{{\sqrt 2}h} (\bx_i - \bx_j) {\1}_{y_i \neq y_j}
\end{align}\end{small}
where ${\tau_1} = \frac{1}{({2{\pi}})^{d/2}{{({\sqrt 2}h)^d}}}$. Moreover,
\begin{small}\begin{align}\label{eq:ise-theorem-seg3}
&\int_{\R^d} {\hat r(\bx,\balpha)} r(\bx) dx \nonumber \\
&=\int_{\R^d} {\hat p(\bx,1)} p(\bx,1) dx + \int_{\R^d} {\hat p(\bx,2)} p(\bx,2) dx
- \int_{\R^d} {\hat p(\bx,1)} p(\bx,2) dx - \int_{\R^d} {\hat p(\bx,2)} p(\bx,1) dx
\end{align}\end{small}
Note that
\begin{small}\begin{align*}
\frac{1}{\tau_0}\int_{\R^d} {\hat p(\bx,1)} p(\bx,1) dx = \sum\limits_{j \colon y_j = 1}  \int_{\R^d} {\balpha_j} K_{h} (\bx - \bx_j) p(\bx,1) dx
\end{align*}\end{small}
We then use the empirical term $\frac{\sum\limits_{i \colon i \neq j} {\balpha_j} K_{h} (\bx_i - \bx_j) {\1}_{y_i = 1}}{n-1}$ to approximate the integral $\int_{\R^d} {\balpha_j} K_{h} (\bx - \bx_j) p(\bx,1) dx$. Since ${\E}_{\{\bx_i,y_i\}_{i \neq j}} \big[ \frac{\sum\limits_{i \colon i \neq j}{\balpha_j}K_{h} (\bx_i - \bx_j) {\1}_{y_i = 1}}{n-1} \big] = \int_{\R^d} {\balpha_j} K_{h} (\bx - \bx_j) p(\bx,1) dx$, and bounded difference holds for $\frac{\sum\limits_{i \colon i \neq j}{\balpha_j} K_{h} (\bx_i - \bx_j) {\1}_{y_i = 1}}{n-1}$, therefore
\begin{small}\begin{align*}
\Pr \Big[ \big|\frac{\sum\limits_{i \colon i \neq j}{\balpha_j} K_{h} (\bx_i - \bx_j) {\1}_{y_i = 1}}{n-1} - \int_{\R^d} {\balpha_j} K_{h} (\bx - \bx_j) p(\bx,1) dx \big| \ge {\balpha_j}\varepsilon \Big] \le 2\exp\big(-2(n-1)\varepsilon^2\big)
\end{align*}\end{small}
And it follows that with probability at least $1-2{n_1}\exp\big(-2(n-1)\varepsilon^2\big)$, where $n_i$ is the number of data points with label $i$,
\begin{small}\begin{align}\label{eq:ise-theorem-seg4}
&\Big | \frac{\sum\limits_{i,j \colon i \neq j, y_i=y_j=1}{\balpha_j} K_{h} (\bx_i - \bx_j) }{n-1} - \frac{1}{\tau_0}\int_{\R^d} {\hat p(\bx,1)} p(\bx,1) dx \Big| \le \sum\limits_{j \colon y_j = 1} \balpha_j \varepsilon
\end{align}\end{small}
Similarly, with probability at least $1-2{n_2}\exp\big(-2(n-1)\varepsilon^2\big)$,
\begin{small}\begin{align}\label{eq:ise-theorem-seg5}
&\Big | \frac{\sum\limits_{i,j \colon i \neq j, y_i=y_j=2}{\balpha_j} K_{h} (\bx_i - \bx_j) }{n-1} - \frac{1}{\tau_0} \int_{\R^d} {\hat p(\bx,2)} p(\bx,2) dx \Big| \le \sum\limits_{j \colon y_j = 2} \balpha_j \varepsilon
\end{align}\end{small}
And it follows from (\ref{eq:ise-theorem-seg4}) and (\ref{eq:ise-theorem-seg5}) that with probability at least $1-2{n}\exp\big(-2(n-1)\varepsilon^2\big)$,
\begin{small}\begin{align}\label{eq:ise-theorem-seg6}
&\Big | \frac{\sum\limits_{i,j \colon i \neq j, y_i=y_j}{\balpha_j} K_{h} (\bx_i - \bx_j) }{n-1} - \frac{1}{\tau_0} \int_{\R^d} \big({\hat p(\bx,1)} p(\bx,1) + {\hat p(\bx,2)} p(\bx,2) \big) dx \Big| \le \varepsilon
\end{align}\end{small}
In the same way, with probability at least $1-2{n}\exp\big(-2n\varepsilon^2\big)$,
\begin{small}\begin{align}\label{eq:ise-theorem-seg7}
&\Big | \frac{\sum\limits_{i,j \colon y_i \neq y_j}{\balpha_j} K_{h} (\bx_i - \bx_j) }{n} - \frac{1}{\tau_0} \int_{\R^d} \big({\hat p(\bx,1)} p(\bx,2) + {\hat p(\bx,2)} p(\bx,1) \big) dx \Big| \le \varepsilon
\end{align}\end{small}
Based on (\ref{eq:ise-theorem-seg6}) and (\ref{eq:ise-theorem-seg7}), with probability at least $1-2{n_2}\exp\big(-2(n-1)\varepsilon^2\big)-2{n}\exp\big(-2n\varepsilon^2\big)$,
\begin{small}\begin{align}\label{eq:ise-theorem-seg8}
&{\rm ISE} (\hat r,r) \le 2 {\tau_0}\frac{\sum\limits_{i,j \colon y_i \neq y_j}{\balpha_j} K_{h} (\bx_i - \bx_j) }{n} - 2{\tau_0} \frac{\sum\limits_{i,j \colon i \neq j, y_i=y_j}{\balpha_j} K_{h} (\bx_i - \bx_j) }{n-1} + \\
&{\tau_1}\sum\limits_{y=1}^2 {\balpha^{(y)}}^{\top} (\bK_{{\sqrt 2}h}){\balpha^{(y)}} -
{\tau_1} \sum\limits_{1 \le i < j \le n} 2{\balpha_i}{\balpha_j} K_{{\sqrt 2}h} (\bx_i - \bx_j) {\1}_{y_i \neq y_j} + 2{\tau_0}\varepsilon \\
& \le 4 {\tau_0}\frac{\sum\limits_{1 \le i < j \le n}(\balpha_i+\balpha_j) K_{h} (\bx_i - \bx_j) {\1}_{y_i \neq y_j} }{n} - {\tau_0} \frac{\sum\limits_{i,j = 1}^ n (\balpha_i+\balpha_j) K_{h} (\bx_i - \bx_j) }{n} + \\
& {\tau_1}\sum\limits_{y=1}^2 {\balpha^{(y)}}^{\top} (\bK_{{\sqrt 2}h}){\balpha^{(y)}} -
{\tau_1} \sum\limits_{1 \le i < j \le n} 2{\balpha_i}{\balpha_j} K_{{\sqrt 2}h} (\bx_i - \bx_j) {\1}_{y_i \neq y_j} + 2{\tau_0}(\frac{1}{n-1}+\varepsilon)
\end{align}\end{small}
And the conclusion of this theorem can be obtained from (\ref{eq:ise-theorem-seg8}).
\end{proof}

Before stating Lemma~\ref{lemma::decompose-similarity}, we introduce the famous spectral theorem in operator theory below.
\begin{MyTheorem}\label{theorem::spectral-theorem}
\textbf{(Spectral Theorem)} Let $L$ be a compact linear operator on a Hilbert space $\cH$.
Then there exists in H an orthonormal basis $\{\phi_1, \phi_2, \ldots\}$ consisting of eigenvectors of $L$. If $\lambda_k$
is the eigenvalue corresponding to $\phi_k$, then the set $\{\lambda_k\}$ is either finite or $\lambda_k \to 0$ when $k \to \infty$.
In addition, the eigenvalues are real if $L$ is self-adjoint.
\end{MyTheorem}

The integral operator by $S$ is defined as
\begin{small}\begin{align*}
&({L_S}f)(\bx) = \int {S(\bx,\bt)f(\bt)} d \bt
\end{align*}\end{small}

\subsection{Proof of Lemma~\ref{lemma::decompose-similarity}}
\begin{proof}
It can be verified that $L_S$ is a compact operator. Therefore, according to Theorem~\ref{theorem::spectral-theorem}, $\{\phi_k\}$ is an orthogonal basis of $\cL^2$. Note that $\phi_k$ is the eigenfunction of $L_S$ with eigenvalue $\lambda_k$ if ${L_S}{\phi_k} = {\lambda_k}{\phi_k}$.

With fixed $\bx \in \cX$, we then have
\begin{small}\begin{align*}
&|\sum\limits_{k = m}^{m+\ell} {\lambda_k}{\phi_k(\bx)}{\phi_k(\bt)}| \le
\big(\sum\limits_{k = m}^{m+\ell} |{\lambda_k}||{\phi_k(\bx)}|^2\big)^{\frac{1}{2}} \cdot \big(\sum\limits_{k = m}^{m+\ell} |{\lambda_k}||{\phi_k(\bt)}|^2\big)^{\frac{1}{2}}
\le  \sqrt{C} \big(\sum\limits_{k = m}^{m+\ell} |{\lambda_k}||{\phi_k(\bx)}|^2\big)^{\frac{1}{2}}
\end{align*}\end{small}
It follows that the series $\sum\limits_{k \ge 1} {\lambda_k}{\phi_k(\bx)}{\phi_k(\bt)}$ converges to a continuous function $e_{\bx}$ uniformly on $\bt$. This is because ${\phi_k} = \frac{{L_S}{\phi_k}}{\lambda_k}$ is continuous for nonzero $\lambda_k$.

On the other hand, for fixed $\bx \in \cX$,  as a function in $\cL^2$,
\begin{small}\begin{align*}
&S(\bx,\cdot) = \sum\limits_{k \ge 1} \langle S(\bx,\cdot), \phi_k\rangle \phi_k = \sum\limits_{k \ge 1} {\lambda_k} \phi_k(\bx) \phi_k(\cdot)
\end{align*}\end{small}

Therefore, for fixed $\bx \in \cX$, $S(\bx,\cdot) = \sum\limits_{k \ge 1} {\lambda_k}{\phi_k(\bx)}{\phi_k(\cdot)} = e_{\bx}(\cdot)$ almost surely w.r.t the Lebesgue measure. Since both are continuous functions, we must have $S(\bx,\bt) = \sum\limits_{k \ge 1} {\lambda_k}{\phi_k(\bx)}{\phi_k(\bt)}$ for any $\bt \in \cX$. It follows that $S(\bx,\bt) = \sum\limits_{k \ge 1} {\lambda_k}{\phi_k(\bx)}{\phi_k(\bt)}$ for any $\bx,\bt \in \cX$.

We now consider two series which correspond to the positive eigenvalues and negative eigenvalues of $L_S$, namely $\sum\limits_{k \colon \lambda_k \ge 0} \lambda_k \phi_k(\bx) \phi_k (\cdot)$ and $\sum\limits_{k:\lambda_k < 0} |\lambda_k| \phi_k(\bx) \phi_k (\cdot)$. Using similar argument, for fixed $\bx$, both series converge to a continuous function, and we let
\begin{small}\begin{align*}
&S^{+}(\bx,\bt) = \sum\limits_{k \colon \lambda_k \ge 0} \lambda_k \phi_k(\bx) \phi_k (\bt) \\
&S^{-}(\bx,\bt) = \sum\limits_{k:\lambda_k < 0} |\lambda_k| \phi_k(\bx) \phi_k (\bt)
\end{align*}\end{small}
$S^{+}(\bx,\bt)$ and $S^{-}(\bx,\bt)$ are continuous function in $\bx$ and $\bt$. All the eigenvalues of $S^{+}$ and $S^{-}$ are nonnegative, and it can be verified that both are PSD kernels since
\begin{small}\begin{align*}
&\sum\limits_{i,j=1}^n c_i c_j S^{+}(\bx_i,\bx_j) = \sum\limits_{i,j=1}^n c_i c_j \sum\limits_{k \colon \lambda_k \ge 0} \lambda_k \phi_k(\bx_i) \phi_k (\bx_j) = \sum\limits_{k \colon \lambda_k \ge 0} \lambda_k   \sum\limits_{i,j=1}^n c_i c_j \phi_k(\bx_i) \phi_k (\bx_j) \\
& = \sum\limits_{k \colon \lambda_k \ge 0} \lambda_k (\sum\limits_{i=1}^n c_i \phi(\bx_i))^2 \ge 0
\end{align*}\end{small}
and similarly for $S^{-}$. Therefore, $S$ is decomposed as $S(\bx,\bt) = S^{+}(\bx,\bt) - S^{-}(\bx,\bt)$.

\end{proof}

\subsection{Proof of Lemma~\ref{lemma::RC-bound-S}}
\begin{proof}
According to Lemma~\ref{lemma::decompose-similarity}, $S$ is decomposed into two PSD kernels as $S = S^{+} - S^{-}$. Therefore, the are two Reproducing Kernel Hilbert Spaces $\cH^{+}$ and $\cH^{-}$ that are associated with $S^{+}$ and $S^{-}$ respectively, and the canonical feature mappings in $\cH^{+}$ and $\cH^{-}$ are $\phi^{+}$ and $\phi^{-}$, with $S^{+}(\bx,\bt) = \langle \phi^{+}(\bx),  \phi^{+}(\bt) \rangle_{H_K^{+}}$ and $S^{-}(\bx,\bt) = \langle \phi^{-}(\bx),  \phi^{-}(\bt) \rangle_{H_K^{-}}$. In the following text, we will omit the subscripts $H_K^{+}$ and $H_K^{-}$ without confusion.

For any $1 \le y \le c$,
\begin{small}\begin{align*}
&h_S(\bx,y) = \sum\limits_{i \colon y_i = y} {\balpha_i}{ S(\bx,{\bx_i})} = \langle \bw^{+}, \phi^{+}(\bx) \rangle - \langle \bw^{-}, \phi^{-}(\bx) \rangle
\end{align*}\end{small}
with $\|\bw^{+}\|^2 = {\balpha^{(y)}}^{\top} {\bS^{+}} {\balpha^{(y)}} \le {B^{+}}^2$ and $\|\bw^{-}\|^2 = {\balpha^{(y)}}^{\top} {\bS^{-}} {\balpha^{(y)}} \le {B^{-}}^2$.
Therefore,
\begin{small}\begin{align*}
&\cH_{S,y} \subseteq \tilde \cH_{S,y} = \{(\bx,y) \to \langle \bw^{+}, \phi^{+}(\bx) \rangle - \langle \bw^{-}, \phi^{-}(\bx) \rangle, \|\bw^{+}\|^2 \le {B^+}^2, \|\bw^{-}\|^2 \le {B^-}^2\}, 1 \le y \le c
\end{align*}\end{small}
and $\cfrakR(\cH_{S,y}) \subseteq \cfrakR(\tilde \cH_{S,y})$. Since we are deriving upper bound for $\cfrakR(\cH_{S,y})$, we slightly abuse the notation and let $\cH_{S,y}$ represent $\tilde \cH_{S,y}$ in the remaining part of this proof.

For $\bx,\bt \in \R^d$ and any $h_S \in \cH_{S,y}$, we have
\begin{small}\begin{align*}
&| h_S(\bx) - h_S(\bt)  | = | \langle \bw^{+}, \phi^{+}(\bx) \rangle - \langle \bw^{-}, \phi^{-}(\bx) \rangle - \langle \bw^{+}, \phi^{+}(\bt) \rangle + \langle \bw^{-}, \phi^{-}(\bt) \rangle | \\
& = |\langle \bw^{+}, \phi^{+}(\bx) - \phi^{+}(\bt)\rangle + \langle \bw^{-}, \phi^{-}(\bt) - \phi^{-}(\bx) \rangle | \\
& \le B^{+} \|\phi^{+}(\bx) - \phi^{+}(\bt)\|  + B^{-} \|\phi^{-}(\bx)-\phi^{-}(\bt)\|) \le (B^{+}+B^{-}) \sqrt{S^{+}(\bx,\bx) + S^{+}(\bt,\bt) + 2 \sqrt{S^{+}(\bx,\bx)S^{+}(\bt,\bt)}   }   \\
&\le 2R^2(B^{+}+B^{-})
\end{align*}\end{small}

We now approximate the Rademacher complexity of the function class $\cH_{S,y}$ with its empirical version $\hat \cfrakR(\cH_{S,y})$ using the sample $\{\bx_i\}$. For each $1 \le y \le c$, Define $E_{\{\bx_i\}}^{(y)} = \hat \cfrakR(\cH_{S,y}) = {\E_{\{\sigma_i\}}}\bigg[\sup_{h_S(\cdot,y) \in \cH_{S,y}} {\Big| \frac{1}{n}\sum\limits_{i=1}^n {\sigma_i} {h_S(\bx_i,y)} \Big| } \bigg]$, then $\sum\limits_{y=1}^c \cfrakR(\cH_{S,y}) = \E_{\{\bx_i\}} \Big[\sum\limits_{y=1}^c E_{\{\bx_i\}}^{(y)}\Big]$, and
\begin{small}\begin{align*}
&\sup_{\bx_1,\ldots, \bx_n, \bx_t^{'}} \Big| E_{\bx_1,\ldots,\bx_{t-1},\bx_t,\bx_{t+1},\ldots,\bx_n}^{(y)} - E_{\bx_1,\ldots,\bx_{t-1},\bx_t^{'},\bx_{t+1},\ldots,\bx_n}^{(y)}\Big| \\
& = \sup_{\bx_1,\ldots, \bx_n, \bx_t^{'}}\Bigg| {\E_{\{\sigma_i\}}}\bigg[\sup_{h_S(\cdot,y) \in \cH_{S,y}} {\Big| \frac{1}{n}\sum\limits_{i=1}^n {\sigma_i} {h_S(\bx_i,y)} \Big| }
-\sup_{h_S(\cdot,y) \in \cH_{S,y}} \Big| \frac{1}{n}\sum\limits_{i \neq t} {\sigma_i} {h_S(\bx_i,y)} + \frac{h_S(\bx_t^{'},y)}{n} \Big|
\bigg] \Bigg| \\
&\le \sup_{\bx_1,\ldots, \bx_n, \bx_t^{'}}{\E_{\{\sigma_i\}}}\Bigg[ \bigg| \sup_{h(\cdot,y) \in \cH_{S,y}} \Big|{\frac{1}{n}\sum\limits_{i=1}^n {\sigma_i} {h_S(\bx_i,y)} \Big| }
-\sup_{h_S(\cdot,y) \in \cH_{S,y}} \Big| \frac{1}{n}\sum\limits_{i \neq t} {\sigma_i} {h_S(\bx_i,y)} + \frac{h_S(\bx_t^{'},y)}{n} \Big| \bigg|
\Bigg]  \\
&\le \sup_{\bx_1,\ldots, \bx_n, \bx_t^{'}}{\E_{\{\sigma_i\}}}\Bigg[ \sup_{h_S(\cdot,y) \in \cH_{S,y}} \bigg|  \Big|{\frac{1}{n}\sum\limits_{i=1}^n {\sigma_i} {h_S(\bx_i,y)} \Big| }
-\Big| \frac{1}{n}\sum\limits_{i \neq t} {\sigma_i} {h_S(\bx_i,y)} + \frac{h_S(\bx_t^{'},y)}{n} \Big| \bigg|
\Bigg] \\
&\le \sup_{\bx_1,\ldots, \bx_n, \bx_t^{'}}{\E_{\{\sigma_i\}}}\Bigg[ \sup_{h_S(\cdot,y) \in \cH_{S,y}} \bigg|  \frac{1}{n}\sum\limits_{i=1}^n {\sigma_i} {h_S(\bx_i,y)}
-\Big( \frac{1}{n}\sum\limits_{i \neq t} {\sigma_i} {h_S(\bx_i,y)} + \frac{h_S(\bx_t^{'},y)}{n} \Big) \bigg|
\Bigg] \\
&= \sup_{\bx_t, \bx_t^{'}} {\E_{\{\sigma_i\}}}\Bigg[ \sup_{h_S(\cdot,y) \in \cH_{S,y}} \bigg|  \frac{h_S(\bx_t,y)}{n} - \frac{h_S(\bx_t^{'},y)}{n}
\bigg|
\Bigg] \le \frac{2R^2(B^{+}+B^{-})}{n}
\end{align*}\end{small}

And it follows that $\Big| \sum\limits_{y=1}^c E_{\bx_1,\ldots,\bx_{t-1},\bx_t,\bx_{t+1},\ldots,\bx_n}^{(y)} - \sum\limits_{y=1}^c E_{\bx_1,\ldots,\bx_{t-1},\bx_t^{'},\bx_{t+1},\ldots,\bx_n}^{(y)}\Big| \le \frac{2R^2(B^{+}+B^{-})c}{n}$. According to the McDiarmid's Inequality,
\begin{small}\begin{align}\label{eq:RC-empirical-S}
\Pr\Big[ \big| \sum\limits_{y=1}^c \hat \cfrakR(\cH_y) - \sum\limits_{y=1}^c \cfrakR(\cH_y)  \big| \ge \varepsilon \Big] \le 2\exp\big(-\frac{n\varepsilon^2}{2(B^{+}+B^{-})^2R^4c^2}\big)
\end{align}\end{small}
Now we derive the upper bound for the empirical Rademacher complexity:

\begin{small}\begin{align}\label{eq:erc-similarity}
&\sum\limits_{y=1}^c \hat \cfrakR(\cH_y) = \sum\limits_{y=1}^c {\E}_{\{\sigma_i\}}\left[\sup_{h_S \in \cH_{S,y}} {|\frac{1}{n} \sum\limits_{i=1}^n \sum\limits_{j=1}^n {\sigma_i}h_S(\bx_i) | } \right] \\
& \le \frac{1}{n} \sum\limits_{y=1}^c {\E}_{\{\sigma_i\}}\left[\sup_{\|\bw^{+}\| \le B^{+}, \|\bw^{-}\| \le B^{-}} {| \sum\limits_{i=1}^n {\sigma_i} \big( \langle  \bw^{+}, \phi^{+}(\bx_i) \rangle  - \langle \bw^{-}, \phi^{-}(\bx_i) \rangle \big)    | } \right] \nonumber \\
& \le \frac{1}{n} \sum\limits_{y=1}^c {\E}_{\{\sigma_i\}}\left[ {B^{+}} \| \sum\limits_{i=1}^n {\sigma_i} \phi^{+}(\bx_i) \| + {B^{-}} \| \sum\limits_{i=1}^n {\sigma_i} \phi^{-}(\bx_i) \|      \right] \nonumber \\
& = \frac{B^{+}c}{n} {\E}_{\{\sigma_i\}}\left[  \| \sum\limits_{i=1}^n {\sigma_i} \phi^{+}(\bx_i) \| \right] +
\frac{B^{-}c}{n} {\E}_{\{\sigma_i\}}\left[ \| \sum\limits_{i=1}^n {\sigma_i} \phi^{-}(\bx_i) \|  \right] \nonumber \\
& \le \frac{B^{+}c}{n} \sqrt{ {\E}_{\{\sigma_i\}}\left[  \| \sum\limits_{i=1}^n {\sigma_i} \phi^{+}(\bx_i) \|^2 \right] } +
\frac{B^{-}c}{n} \sqrt{ {\E}_{\{\sigma_i\}}\left[  \| \sum\limits_{i=1}^n {\sigma_i} \phi^{-}(\bx_i) \|^2 \right] } \nonumber \\
& \le \frac{B^{+}c}{n} \sqrt{ \sum\limits_{i}^n s^{+}(\bx_i,\bx_i) } +
\frac{B^{-}c}{n} \sqrt{  \sum\limits_{i}^n s^{-}(\bx_i,\bx_i) } \le \frac{Rc}{\sqrt{n}} (B^{+} + B^{-}) \nonumber
\end{align}\end{small}

Also, by (\ref{eq:kernel-error-seg4}) in the proof of Lemma~\ref{lemma::RC-bound}, (\ref{eq:RC-empirical-S}) and (\ref{eq:erc-similarity}), with probability at least $1-\delta$, we have
\begin{small}\begin{align}\label{eq:kernel-error-seg6}
&\cfrakR(\cH_S) \le (2c-1)\sum\limits_{y=1}^c \cfrakR(\cH_y) \nonumber \\
&\le \frac{R(2c-1){c}(B^{+} + B^{-})}{\sqrt n} + 2c(2c-1)(B^{+}+B^{-})R^2\sqrt{\frac{\ln{\frac{2}{\delta}}}{2n}}
\end{align}\end{small}

\end{proof}

\subsection{Similarity Machines: Learning Max-Margin Classifier with General Similarity Function}
In this section, we introduce Similarity Machines as a generalization of Kernel SVMs, which is a framework of learning maximum margin classifier with general similarity matrix $S$ which is symmetric but not necessarily positive semi-definite. We consider binary classification in this section, and it can be extended to the case of multi-class in a way similar to multi-class SVMs.

\subsubsection{Notations}
Suppose the data $\cS = \{\bx_i, y_i\}_{i=1}^n$ are generated i.i.d. from some distribution supported on $\cX \times \cY$, $\bx_i \in \cX$, $y_i \in \cY = \{\pm 1\}$ is the label of $\bx_i$. 


\subsubsection{Rademacher Complexity for Kernel SVMs}
A kernel $k \colon \cX \times \cX \to \R$ is a continuous function such that for all $\{\bx_i\}_{i=1}^n$, the Gram matrix $K$, with $K_{ij} = k(\bx_i, \bx_j)$, is positive semi-definite and symmetric. Kernel SVMs uses the following kernel expansions as the classifier:
\begin{small}\begin{align}\label{eq:kernel-svm-classifier}
&\bx \to \sum\limits_{i=1}^n {\alpha_i}k(\bx,\bx_i)
\end{align}\end{small}
The kernel $k$ is associated with a feature map $\Phi \colon \cX \to \cH$ where $\cH$ is a Hilbert space with inner product $\langle \cdot, \cdot \rangle$, and $k(\bx,\bx') = \langle \Phi(\bx), \Phi(\bx') \rangle$ for all $\bx, \bx' \in \cX$. Denote by $\|\cdot\|$ the norm in $\cH$, then
\begin{small}\begin{align*}
& \|\sum\limits_{i=1}^n {\alpha_i}\Phi(\bx_i) \|^2 = \sum\limits_{i,j=1}^n {\alpha_i}{\alpha_j}k(\bx_i,\bx_j)
\end{align*}\end{small}
And the function class $F$ that the classifier (\ref{eq:kernel-svm-classifier}) belongs to is 
\begin{small}\begin{align*}
&F = \{ f(\bx) = \sum\limits_{i=1}^n {\alpha_i}k(\bx,\bx_i) \colon \bx, \bx_i \in \cX, \sum\limits_{i,j=1}^n {\alpha_i}{\alpha_j}k(\bx_i,\bx_j) \le B^2 \} \subseteq \{f(\bx) = \langle \bw, \Phi(\bx) \rangle \colon \bw \in \cH, \|\bw\| \le B \}
\end{align*}\end{small}
The empirical Rademacher complexity of the function class $F$ satisfies
\begin{small}\begin{align}\label{eq:erc-kernel}
&\hat \cfrakR(F) = {\E}_{\{\sigma_i\}}\left[\sup_{f \in F} {|\frac{1}{n} \sum\limits_{i=1}^n {\sigma_i}{f(\bx_i)} | }  \right] \\
& \le {\E}_{\{\sigma_i\}}\left[\sup_{\bw \le B} {|\frac{1}{n} \sum\limits_{i=1}^n {\sigma_i}{\langle \bw, \Phi(\bx_i) \rangle} | }  \right] \nonumber \\
& \le \frac{B}{n} {\E}_{\{\sigma_i\}}\left[ { \|\sum\limits_{i=1}^n {\sigma_i}{\Phi(\bx_i)} \| }  \right] \nonumber \\
& = \frac{B}{n} {\E}_{\{\sigma_i\}}\left[ { \sqrt { \|\sum\limits_{i=1}^n {\sigma_i}{\Phi(\bx_i)} \|^2 } }  \right] \nonumber \\
& \le \frac{B}{n}  \Big( {\E}_{\{\sigma_i\}}\left[ { \sum\limits_{i,j=1}^n {\sigma_i}{\sigma_j}{k(\bx_i,\bx_j)}  }  \right] \Big)^{\frac{1}{2}}
 = \frac{B}{n} \sqrt{ \sum\limits_{i}^n k(\bx_i,\bx_i) }
\end{align}\end{small}
The following theorem in \citep{Bartlett2003} gives a margin-based estimate of misclassification probability, namely, the classification error, for any function in $F$ when it is used as the classifier.
\begin{MyTheorem}[Theorem $21$ in \citep{Bartlett2003}]
\label{theorem::kernel-svm-bound}
Fix $B,\gamma>0$, let $k \colon \cX \times \cX \to \R$ be a kernel with $\sup_{\bx \in \R} k(\bx,\bx) < \infty$. Define the margin cost function $\phi$ as
\begin{small}\begin{align*}
&\varphi(t) =
\left\{
\begin{array}
    {r@{\quad:\quad}l}
    1 & {t \le 0 } \\
    1-\frac{t}{\gamma} & {0 < t < \gamma} \\
    0 & {t \ge \gamma}
\end{array}
\right.
\end{align*}\end{small}
Suppose $\cS = \{\bx_i, y_i\}_{i=1}^n$ are generated i.i.d. from some distribution $P$ supported on $\cX \times \{ \pm 1 \}$, then with probability at least $1-\delta$, every function $f$ of the form
\begin{small}\begin{align*}
&f(\bx) = \sum\limits_{i=1}^n {\alpha_i}k(\bx,\bx_i)
\end{align*}\end{small}
with $\sum\limits_{ij} {\alpha_i}{\alpha_j} k(\bx_i,\bx_j) \le B^2$ satisfies
\begin{small}\begin{align}\label{eq:kernel-svm-bound}
&\Pr[\by f(\bx) \le 0] \le \hat \E_n [\varphi(\by f(\bx))] + \frac{4B}{ n\gamma} \sqrt{\sum\limits_{i=1}^n k(\bx_i,\bx_i)} + \Big(\frac{8}{\gamma}+1\Big) \sqrt{\frac{\log 4/{\delta}}{2n}}
\end{align}\end{small}
\end{MyTheorem}

\subsubsection{Rademacher Complexity of Similarity Machines}
We now consider Similarity Machines which uses a general continuous function $s \colon \cX \times \cX \to \R$ as a classifier, which is not necessarily a positive semi-definite kernel. When $s$ satisfies the assumptions in Lemma~\ref{lemma::decompose-similarity}, it has decomposition
\begin{small}\begin{align*}
&s(\bx,\bt) =  s^{+}(\bx,\bt) - s^{-}(\bx,\bt)
\end{align*}\end{small}%

The function that is used as the classifier by Similarity Machines has the form 
\begin{small}\begin{align}\label{eq:hypothesis}
&h(\bx) = \sum\limits_{i} {\alpha_i}{s(\bx,{\bx_i})}
\end{align}\end{small}



Suppose the feature map associated with $s^{+}$ and $s^{-}$ are $\Phi^{+}$ and $\Phi^{-}$ and the associated RKHS are $\cH_1$ and $\cH_2$. Let $\sum\limits_{i,j=1}^n {\alpha_i}{\alpha_j} s^{+}(\bx_i,\bx_j) \le B_1^2$ and $\sum\limits_{i,j=1}^n {\alpha_i}{\alpha_j} s^{-}(\bx_i,\bx_j) \le B_2^2$, then the function class $\cH_s$ that the function $h(\bx)$ (\ref{eq:hypothesis}) belongs to is
\begin{small}\begin{align*}
&\cH_s = \{ f_s(\bx) = \sum\limits_{i=1}^n {\alpha_i}s(\bx,\bx_i) \colon \bx, \bx_i \in \cX, \sum\limits_{i,j=1}^n {\alpha_i}{\alpha_j} s^{+}(\bx_i,\bx_j) \le B_1^2, \sum\limits_{i,j=1}^n {\alpha_i}{\alpha_j} s^{-}(\bx_i,\bx_j) \le B_2^2 \} \\
&\subseteq
\{h(\bx,y) = \langle \bw^{+}, \Phi^{+}(\bx) \rangle - \langle \bw^{-}, \Phi^{-}(\bx) \rangle \colon \bw^{+} \in \cH_1, \bw^{-} \in \cH_2,\|\bw^{+}\| \le B_1, \|\bw^{-}\| \le B_2 \}
\end{align*}\end{small}%

The empirical Rademacher complexity of $\cH_s$ satisfies
\begin{small}\begin{align}\label{eq:erc-similarity}
&\hat \cfrakR(\cH_s) = {\E}_{\{\sigma_i\}}\left[\sup_{h \in \cA} {|\frac{1}{n} \sum\limits_{i=1}^n \sum\limits_{j=1}^n {\sigma_i}h(\bx_i) | } \right] \\
& \le \frac{1}{n}{\E}_{\{\sigma_i\}}\left[\sup_{\|\bw^{+}\| \le B_1, \|\bw^{-}\| \le B_2} {| \sum\limits_{i=1}^n {\sigma_i} \big( \langle  \bw^{+}, \Phi^{+}(\bx_i) \rangle  - \langle \bw^{-}, \Phi^{-}(\bx_i) \rangle \big)    | } \right] \nonumber \\
& \le \frac{1}{n}{\E}_{\{\sigma_i\}}\left[ {B_1} \| \sum\limits_{i=1}^n {\sigma_i} \Phi^{+}(\bx_i) \| + {B_2} \| \sum\limits_{i=1}^n {\sigma_i} \Phi^{-}(\bx_i) \|      \right] \nonumber \\
& = \frac{B_1}{n} {\E}_{\{\sigma_i\}}\left[  \| \sum\limits_{i=1}^n {\sigma_i} \Phi^{+}(\bx_i) \| \right] +
\frac{B_2}{n} {\E}_{\{\sigma_i\}}\left[ \| \sum\limits_{i=1}^n {\sigma_i} \Phi^{-}(\bx_i) \|      \right] \nonumber \\
& \le \frac{B_1}{n} \sqrt{ \sum\limits_{i}^n s^{+}(\bx_i,\bx_i) } +
\frac{B_2}{n} \sqrt{ \sum\limits_{i}^n s^{-}(\bx_i,\bx_i) } \nonumber
\end{align}\end{small}%

We can observe that when $s$ is a positive semi-definite kernel, then $s^{-}  \equiv 0 $ and the Rademacher complexity bound for function class with general continuous function coincides with that for function class with positive semi-definite kernel in (\ref{eq:erc-kernel}). Applying Theorem $7$ and Theorem $11$ in \citep{Bartlett2003}, we have the following theorem establishing the generalization error bound for Similarity Machines:

\begin{MyTheorem}[Generalization Error Bound for Similarity Machines]
\label{theorem::similarity-machine-bound}
Fix $B,\gamma>0$, let $k \colon \cX \times \cX \to \R$ be a kernel with $\sup_{\bx \in \R} k(\bx,\bx) < \infty$. Define the margin cost function $\phi$ as
\begin{small}\begin{align*}
&\varphi(t) =
\left\{
\begin{array}
    {r@{\quad:\quad}l}
    1 & {t \le 0 } \\
    1-\frac{t}{\gamma} & {0 < t < \gamma} \\
    0 & {t \ge \gamma}
\end{array}
\right.
\end{align*}\end{small}
Suppose $\cS = \{\bx_i, y_i\}_{i=1}^n$ are generated i.i.d. from some distribution $P$ supported on $\cX \times \{ \pm 1 \}$, then with probability at least $1-\delta$, every function $f$ of the form
\begin{small}\begin{align*}
&f(\bx) = \sum\limits_{i=1}^n {\alpha_i}s(\bx,\bx_i)
\end{align*}\end{small}%
where $s$ satisfies the assumptions in Lemma~\ref{lemma::decompose-similarity} with decomposition $s(\bx,\bt) =  s^{+}(\bx,\bt) - s^{-}(\bx,\bt)$, and $\sum\limits_{i,j=1}^n {\alpha_i}{\alpha_j} s^{+}(\bx_i,\bx_j) \le B_1^2$ and $\sum\limits_{i,j=1}^n {\alpha_i}{\alpha_j} s^{-}(\bx_i,\bx_j) \le B_2^2$. Then, 
\begin{small}\begin{align}\label{eq:similarity-machine-bound}
&\Pr[\by f(\bx) \le 0] \le \hat \E_n [\varphi(\by f(\bx))] + \frac{4}{ n\gamma} ({B_1} \sqrt{ \sum\limits_{i}^n s^{+}(\bx_i,\bx_i) } +
{B_2} \sqrt{ \sum\limits_{i}^n s^{-}(\bx_i,\bx_i) }) + \Big(\frac{8}{\gamma}+1\Big) \sqrt{\frac{\log 4/{\delta}}{2n}}
\end{align}\end{small}
\end{MyTheorem}

It can also be observed that when $s$ is a positive semi-definite kernel, the error bound for Similarity Machines (\ref{eq:similarity-machine-bound}) reduces to that for Kernel SVMs (\ref{eq:kernel-svm-bound}).

\end{document}